\documentclass[lettersize,journal]{IEEEtran}
\usepackage{amsmath,amsfonts}
\usepackage{algorithmic}
\usepackage{algorithm}
\usepackage{array}
\usepackage[caption=false,font=normalsize,labelfont=sf,textfont=sf]{subfig}
\usepackage{textcomp}
\usepackage{stfloats}
\usepackage{url}
\usepackage{verbatim}
\usepackage{graphicx}
\usepackage{cite}

\usepackage{multicol}
\usepackage{float}
\usepackage{xcolor}

\usepackage[font=small,labelfont=bf]{caption}
\usepackage{multirow}
\usepackage{graphicx}
\usepackage{grffile}
\usepackage{enumitem}
\usepackage{makecell}

\usepackage{amsthm}
\newtheorem{theorem}{Theorem}[section]

\newtheorem{Proposition}[theorem]{Proposition}

\newcommand{\revisionPQ}[2]{{\color[rgb]{0,0,0}#2}}

\usepackage{pifont}

\newcommand{\cmark}{\ding{51}}%
\newcommand{\xmark}{\ding{55}}%

\usepackage{adjustbox}
\usepackage{booktabs}
\begin{document}

\title{On the amplification of security and privacy risks by post-hoc explanations in machine learning models}


\author{Pengrui~Quan\IEEEauthorrefmark{1},
        Supriyo Chakraborty\IEEEauthorrefmark{2},
        Jeya Vikranth Jeyakumar\IEEEauthorrefmark{1},
        and~Mani~Srivastava\IEEEauthorrefmark{1}
\thanks{Pengrui Quan is with the Department
of Electrical and Computer Engineering, University of California, Los Angeles, California,
CA, 90095 USA e-mail: prquan@g.ucla.edu.}\\
\IEEEauthorblockA{\IEEEauthorrefmark{1}University of California, Los Angeles}\\
\IEEEauthorblockA{\IEEEauthorrefmark{2}IBM Thomas J. Watson Research Center}
}

\markboth{Journal of \LaTeX\ Class Files,~Vol.~14, No.~8, August~2021}%
{Shell \MakeLowercase{\textit{et al.}}: A Sample Article Using IEEEtran.cls for IEEE Journals}


\maketitle


\begin{abstract}
    A variety of explanation methods have been proposed in recent years to help users gain insights into the results returned by neural networks, which are otherwise complex and opaque black-boxes. However, explanations give rise to potential side-channels that can be leveraged by an adversary for mounting attacks on the system. In particular, \revisionPQ{}{post-hoc} explanation methods that highlight input dimensions according to their importance or relevance to the result also leak  information that weaken security and privacy. In this work, we perform the first systematic characterization of the privacy and security risks arising from various popular explanation techniques. First, we propose novel explanation-guided black-box evasion attacks that lead to $10\times$ reduction in query count for the same success rate. We show that the adversarial advantage from explanations can be quantified as a \revisionPQ{}{reduction in the total variance of the estimated gradient}. Second, we revisit the membership information leaked by common explanations. Contrary to observations in prior studies, via our modified attacks we show significant leakage of membership information (above $100\%$ improvement over prior results), even in a much stricter black-box setting. Finally, we study explanation-guided model extraction attacks and demonstrate adversarial gains through large reduction in query count. 
\end{abstract}


\section{Introduction}
\label{sec:intro}

Post-hoc explanations~\cite{simonyan2013deep, springenberg2014striving, smilkov2017smoothgrad, sundararajan2017axiomatic} have emerged as the primary means of introducing decisional transparency into deep neural networks-based systems, which are otherwise complex and operate as opaque black-boxes. These explanations are usually presented to the user in the form of a saliency map over the input, highlighting features that were the most relevant to the model decision. Thus, deployed models output both their decision as well as the explanation of the decision to ensure reliability~\cite{tonekaboni2019clinicians}, fairness~\cite{anders2020fairwashing}, and compliance to data regulatory policies \footnote{\url{https://gdpr-info.eu/}}.

However, as these post-hoc explanations are derived using information from the model's parameters and gradients, they also serve as an information-rich side-channel that is available to an adversary, in addition to the model decision. For instance, ~\cite{shokri2021privacy, milli2019model} have shown that explanations can leak private information about both the model and data attributes. \revisionPQ{}{In this paper, we perform a systematic characterization of the privacy and security risks, quantified by \emph{increased efficiency} (e.g., reduced query count, attack feasibility in constrained settings and so on) of evasion attacks (EA), membership inference attacks (MIA) and model extraction attacks (MEA), due to the release of explanations.}

We consider evasion attacks on real-world systems, which are usually mounted in a \emph{black-box setting}. An adversary (with no access to the model parameters and gradients) repeatedly queries the victim model (that is being attacked) to design adversarial examples, eliciting targeted misclassification~\cite{ilyas2018black, chen2017zoo}. However, the cumulative costs associated with a large number of queries can render a query-inefficient black-box attack impractical~\cite{ilyas2018black}. Towards this end, we propose three novel explanation-guided evasion attacks, that significantly reduce the number of queries needed, making these attacks highly feasible in practice. In (\romannumeral 1) Explanation-Guided Transfer Attack (EGTA), we directly use  explanations as a transfer prior to shape a search distribution for estimating the gradient~\cite{ilyas2018black}. This attack is highly efficient when the estimated gradient (using explanations) has high cosine similarity to the true gradient; (\romannumeral 2) If the cosine similarity is low, using the explanation as a prior leads to a highly biased gradient estimation. The Explanation-Guided Surrogate Model Attack (EGSMA), instead uses explanations to train a surrogate model and mimic the decisions of the victim model. Gradient from the surrogate model is then used to generate the adversarial examples. (\romannumeral 3) Finally, when absolute value of the true gradients is used to generate explanations~\cite{smilkov2017smoothgrad, sundararajan2017axiomatic}, we propose to use the Explanation-Guided Sampling Attack (EGSA). In this, we use explanations to shape the covariance matrix of the search distribution, used to estimate the gradients. Finally, we show that the adversarial advantage due to explanations can be quantified as a reduction in the total variance of the estimated gradients.

We then revisit the problem of membership information leaked by explanations. In this, the goal of an adversary is to exploit model explanations to infer the membership of a sample in the training data. We build on the intuition provided in prior work that variance in explanations can be used as an indicator of membership, in black-box settings~\cite{shokri2021privacy}. However, we use the softmax score of the output, as opposed to the logit layer score (as proposed earlier) to generate the explanation and compute the variance. Contrary to the observations in \cite{shokri2021privacy}, using our modified attack, we found significant leakage of membership information via explanations.

Finally, we study the adversarial advantage due to explanations in mounting model extraction attacks. The goal of the adversary is to train a surrogate model that is \emph{functionally equivalent} to the victim model in a black-box setting. We demonstrate significant query reduction and also connect the adversarial gain to the mean square values (MSV) of explanations: a higher MSV results in a larger adversarial advantage in performing MEA. 

\noindent{\bf Choice of Explanation Techniques:} To determine a representative subset of explanation techniques, we use three metrics: (i) \textit{Fidelity}~\cite{yeh2019fidelity} which focuses on the closeness of the predictions between the true model and the linear model approximated by the explanations given random input perturbation; (ii) \textit{ROAR}~\cite{hooker2019benchmark} computes the change in model accuracy after removing the important features and retraining; (iii) \textit{User study}~\cite{jeyakumar2020can} identifies explanation types that are preferred by end-users. These can differ based on the modality of the data that are used to evaluate and rank explanations. Accordingly after a survey of various papers, we selected six explanation techniques (Table \ref{table:criteria}) which ranked highly as per the above metrics.

\begin{table}[]
\centering
\begin{adjustbox}{max width=0.40\textwidth}

    \begin{tabular}{l|l}
    \hline
    Explanation & criterion\\ 
    \hline
    Gradient~\cite{simonyan2013deep} & Fidelity~\cite{yeh2019fidelity}, ROAR~\cite{hooker2019benchmark}\\
    \hline
    Guided backprop~\cite{springenberg2014striving} & ROAR~\cite{hooker2019benchmark}\\
    \hline
    Smooth Grad~\cite{smilkov2017smoothgrad} & Fidelity~\cite{yeh2019fidelity}\\
    \hline
    Int. Grad~\cite{sundararajan2017axiomatic} & Fidelity~\cite{yeh2019fidelity}\\
    \hline
    Grad-Cam~\cite{selvaraju2017grad} & User study~\cite{jeyakumar2020can}\\
    \hline
    LIME~\cite{ribeiro2016should} & User study~\cite{jeyakumar2020can}\\
    \hline
    \end{tabular}
\end{adjustbox}
\caption{Explanations and their corresponding metrics}\label{table:criteria}
\vspace{-2em}
\end{table}

In summary, we make the following contributions:

\begin{itemize}
    \item We exploit explanations as a side-channel and develop explanation-guided black box evasion attacks. We show a $10\times$ reduction in query count when a gradient-based explanation is available. We provide theoretical analysis of the adversarial gain and quantify it as the reduction in variance of the gradient estimate.
    
    \item We propose a modified attack for extracting membership information from explanations and report significant leakage compared to prior work.
    
    \item We demonstrate large query reduction for model extraction attack using explanations.
\end{itemize}
\section{Related Work}
\label{sec:related}
We group prior work into two categories: (i) Post-hoc explanation mechanisms; and (ii) The intersection of explanation and security and privacy risks.\\

\noindent{\bf Post-hoc explanation mechanisms:} These are after-the-prediction explanation mechanisms that aim to provide evidence to elucidate the model decision~\cite{expl_survey,jeyakumar2020can}. Post-hoc explanation techniques are broadly divided into (i) textual explanations of system outputs, (ii) visualisations of learned representations or models (e.g., saliency maps), and (iii) explanations-by-example~\cite{jeyakumar2020can}. In our work, we focus on (ii) post-hoc mechanisms that generate saliency maps, and describe them briefly as follows.

\begin{itemize}[leftmargin=*]
\item Gradient-based saliency map: GradientInput~\cite{simonyan2013deep}, IntegratedGradient~\cite{sundararajan2017axiomatic}, GradCam~\cite{selvaraju2017grad}, and SmoothGrad~\cite{smilkov2017smoothgrad} visually highlight the feature components that has large affects on the model prediction using gradient-based techniques.
\item Local interpretable surrogate model: LIME~\cite{ribeiro2016should} transforms a black-box model into a transparent one by training a surrogate interpretable model on the point of interests.

\end{itemize}

\medskip
\noindent{\bf Model Explanations and Adversarial Attacks:} Complementary to our work, there has been several recent papers that have explored the connection between adversarial attacks and explanations. 

\begin{itemize}[leftmargin=*]

\item Explanation-Guided Membership Inference Attack: \cite{shokri2021privacy} has shown that an adversary can exploit model explanations to infer whether a data belongs to the training set of a model, i.e., mounting Membership Inference Attack (MIA).

\item Explanation-Guided Model extraction Attack: Besides, \cite{milli2019model} demonstrate that gradient-based explanations of a model can help the adversary to perform efficient Model Extraction Attack (MEA), i.e., copying the model without authorization, contradicting the intention of keep a proprietary model secret.

\end{itemize}
Orthogonal to our study, there was also been work on attacking explanations for various objectives such as reducing trust on the model decision~\cite{iccv_fooling} and fairwashing~\cite{anders2020fairwashing}.

However, none of the above study how explanations can benefit evasion attacks. Besides, there has not been a systematic study on the security and privacy risk for popular explanations types. Therefore, this work aims at filling the aforementioned gaps.

\section{Evasion Attacks (EA)}
\label{sec:problem}
\begin{table}[]
\begin{adjustbox}{max width=0.45\textwidth}
    \begin{tabular}{l|l}
    \hline
    Notation & Definition\\ 
    \hline
    ${\bf x}$ & Input to the model\\
    \hline
    ${F(\bf x)}_i$ & Logit score of the $i$-th class\\
    \hline
    $f(\bf x)=\boldsymbol{\sigma}(F(\bf x))$ & Softmax scores of the output logit \\
    \hline
    $\nabla_{\mathbf x} f(\mathbf x)_i$ & Gradient of the softmax score of the $i$-th class w.r.t input\\
    \hline
    $E(F, \bf x)$ & Explanation in the input space produced by the logit\\ 
    \hline
    $E(f, \bf x)$ & Explanation produced by the softmax score\\ 
    \hline
    $\boldsymbol{e}$ & An instance produced by $E$ (true explanation)\\ 
    \hline
    $|\boldsymbol{e}|$ & \makecell{Element-wise absolute value of $\boldsymbol{e}$ \\
    (absolute-value explanation)}\\ 
    \hline
    \end{tabular}
\end{adjustbox}
\caption{Summary of notations used in the paper}\label{table:notations}
\vspace{-2em}
\end{table}

{\bf Notation:} Let ${\bf x} \in \mathbb{R}^n$ be the input and ${\bf x}_{adv} \in \mathbb{R}^n$ be the corresponding adversarial example that we want to generate. We denote the victim model as $F(\bf{x}): \mathbb{R}^n \rightarrow \mathbb{R}^k$ where $k$ is the number of classes and $f(\bf x)$ is the corresponding probability for each class. Let $\nabla_{\mathbf x} f(\mathbf x)_i \in \mathbb{R}^{n}$ be the gradient of the output probability (softmax score) w.r.t the input for the $i-$th label. We denote by $E: \mathbb{R}^{n} \rightarrow \mathbb{R}^n$ the class of explanation mechanisms, that take gradient vector for an output label as input and generates a saliency map as the output. The saliency map is the same size as the input. Table. \ref{table:notations} summarizes the common notations used.

\noindent{\bf Adversary Access:} In most practical settings, adversarial examples must be considered under a limited threat model which prompted us to study black-box attacks, where the adversary only has query-access to the target model~\cite{ilyas2018black}. 
In this setting, an adversary provides a query input $\mathbf x$ to the model and can only observe the model's response to the query. Specifically, the response consists of the output of the explanation mechanism in addition to the output of the victim model. The adversary does not have access to the model weights and cannot directly compute the gradient w.r.t the model parameters.

\subsection{Background}
We consider $l_2$-norm bounded perturbation of the input with the goal of performing both targeted (i.e., misclassification to a given target class) and untargeted (i.e., misclassification to any class other than the true class) attacks. Based on additional constraints on the model response in the black-box setting, we consider attacks under the following two threat models:

\subsubsection{Soft-label attack} 
In this setting, the adversary has access to the predicted probabilities over all or the top-k classes. Prior work have used the information to generate adversarial inputs via zeroth-order optimization methods (e.g., finite-difference method)~\cite{chen2017zoo, ilyas2018black} and gradient-free techniques (e.g., genetic algorithm)~\cite{alzantot2019genattack}. Here an adversary has access to the predicted class $t$, as well as the prediction probability vector $f({\bf x})_y$ for all classes $y$. We define a loss function $\mathcal{L}({\mathbf x})$ as $\mathcal{L}(\mathbf{x}) = f(\mathbf{x})_t\mathbf{1}_{\{f(\mathbf{x})_t>\underset{i\neq t}{\mathrm{max}}\text{ }f(\mathbf{x})_i \}}$.
Our objective in this setting is to perform an untargeted attack. Note, $\mathcal{L}({\mathbf x})=0$ only when the predicted class does not equal the target class $t$. The attack can thus be formulated as $\mathbf{x}_{adv} = \underset{\mathbf{z}: ||\mathbf{x}-\mathbf{z}||_2\leq \epsilon}{\mathrm{arg\,min}} \mathcal{L}(\mathbf{z})$, where $\epsilon$ is the perturbation budget.
The above objective can be optimized by taking a step in the direction of the gradient of the loss followed by a projection onto the feasible set. We follow the modified NES gradient-free approach~\cite{wierstra2014natural} to estimate the gradient of $f(\mathbf{x})$ as:
\begin{equation}\label{nes}
    \widetilde{\nabla f(\mathbf{x})} = \frac{1}{2\delta B} \sum_{b=1}^B (f(\mathbf{x}+\delta\mathbf{u}_b)_i-f(\mathbf{x}-\delta\mathbf{u}_b)_i )\mathbf{u}_b 
\end{equation}
where $\delta$ is the perturbation parameter, $\mathbf{u}_b \in \mathbb{R}^n$, and $\{\mathbf{u}_b\}_{b=1}^{B}$ are i.i.d. samples from the normal distribution $\mathcal N(\mathbf{0}, \epsilon\mathbf{I})$, in a manner similar to \cite{wierstra2014natural, ilyas2018black}. Given the estimated gradient, the adversarial inputs are generated via an approach similar to PGD~\cite{madry2017towards}.

\subsubsection{Hard-label attack}
In this setting, only the top-1 predicted label is output by the model ~\cite{chen2020hopskipjumpattack, cheng2019sign}. The resulting discontinuous and combinatorial attack problem is first transformed into a random-walk problem along the decision boundary in~\cite{chen2020hopskipjumpattack}. Below we summarize the formulation of the hard-label attack for generating an adversarial example as proposed in~\cite{chen2020hopskipjumpattack}.
We define a decision function $\phi_t(\mathbf{x})$ as:
\begin{equation}\label{M}
	\begin{aligned}
	& &\phi_t(\mathbf{x}) \overset{def}{:=} \left\{\begin{array}{ll}
	1, & \text{if prediction is in the target class } t\\
	-1, &  \text{otherwise}\\
	\end{array}
	\right.\\
	\\
	\end{aligned}
\end{equation}
A decision attack objective can then be formulated as $\mathbf{x}_{adv} = \underset{\mathbf{z}: \phi_t(\mathbf{z})=1}{\mathrm{arg\,min}}||\mathbf{x}-\mathbf{z}||$.
The goal is to minimize the distance of the adversarial sample to $\bf x$ while satisfying the boundary constraint $\phi_t(\mathbf{x})=1$. This is performed via a random walk along the decision boundary, in which the direction of the next step is estimated using the gradient of $\phi_t(\mathbf{x})$ as:
\begin{equation}\label{hsja}
    \widetilde{\nabla \phi_t}(\mathbf{x}) = \frac{1}{B}\sum_{b=1}^B \phi_t(\mathbf{x}+\delta \mathbf{u}_b) \mathbf{u}_b
\end{equation}

where $\delta$ is the perturbation parameter, $\mathbf{u}_b \in \mathbb{R}^n$, and $\{\mathbf{u}_b\}_{b=1}^{B}$ are i.i.d. samples from the normal distribution $\mathcal N(\mathbf{0}, \epsilon\mathbf{I})$.




\subsection{Our Attack Strategies for EA}
\label{sec:strategies}
Generating adversarial examples in the black-box setting requires an adversary to estimate the model gradient. Rather than optimizing for a specific attack objective 
prior work have explored the idea of a \emph{search distribution} and have established that an adversary can maximize the expected value of the attack objective, under the search distribution, using far fewer number of queries~\cite{ilyas2018black}.

In our attacks we adopt this idea of sampling from a suitable search distribution to estimate the gradient vector. We propose three different attacks, each designed to exploit the information in model explanation in different ways. The attacks are applicable for soft-label and hard-label settings and can be used for both targeted, and untargeted misclassification.



\begin{figure}[h]
\includegraphics[width=0.98\linewidth]{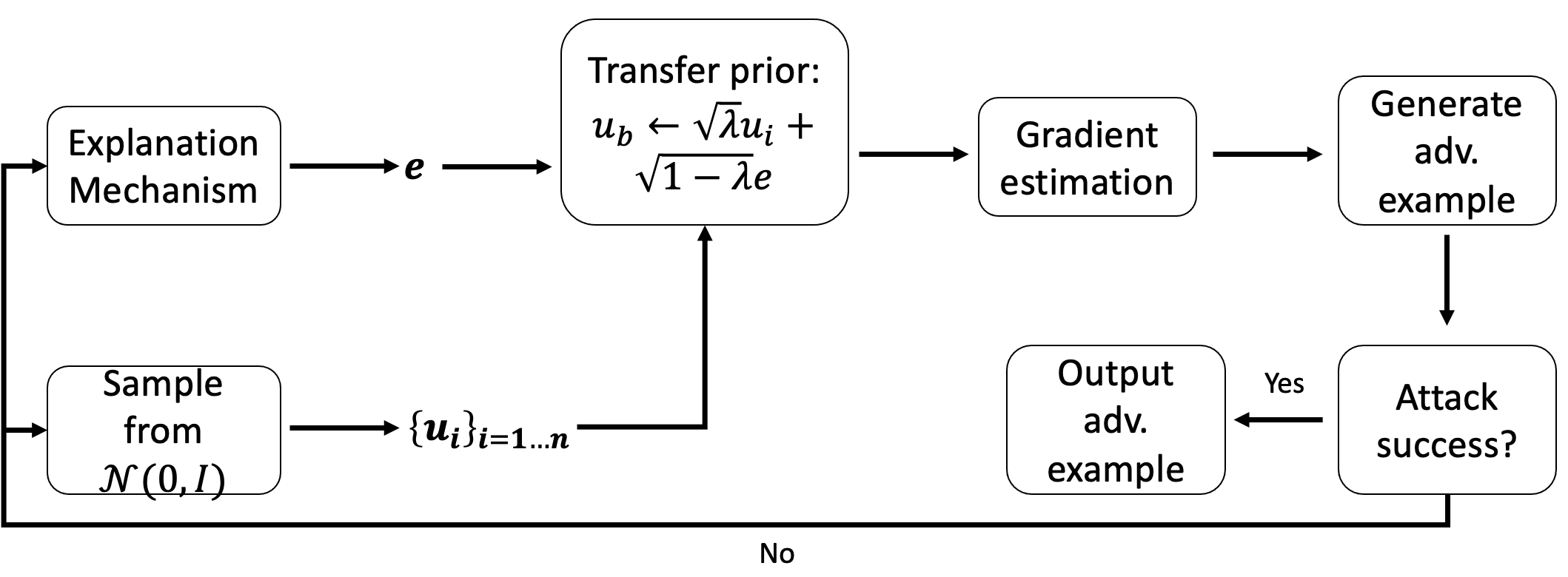}  \centering
	\caption{Explanation-Guided Transfer Attack  (EGTA)}
	\label{fig:egta}
\end{figure}
\noindent{\textbf{Explanation-guided Transfer Attack (EGTA)}} We consider an adversary that samples vectors $\mathbf{u}_b$ from a normal distribution $\mathcal N(\mathbf{0}, \epsilon \mathbf{I})$ to perform gradient estimation. If the adversary had prior knowledge that gradient ${\bf g} = \nabla_{\bf x} F({\bf x})$ is used to compute the explanation ${\bf e} = E(F, \bf{x})$ then it can use the explanation to shape the search distribution by incorporating it as a prior. Besides, we also let its norm $||\boldsymbol e||=1$. In this case, the sampled vector $\mathbf{u}_b \in \mathbb{R}^n$ for  \eqref{nes} and \eqref{hsja} can be expressed as:

\begin{equation}\label{eq:direct_transfer}
    \mathbf{u}_b = \sqrt{\lambda} \mathbf{u} + \sqrt{1-\lambda} \mathbf{e}
\end{equation}
where $\mathbf{u} \sim \mathcal N(\mathbf{0}, \epsilon\mathbf{I})$. We use $\lambda \in [0, 1]$ to reflect adversarial confidence of the prior knowledge. We adopt \eqref{eq:direct_transfer} as our attack strategy and name it as Explanation-guided Transfer Attack (EGTA) as shown in Fig. \ref{fig:egta}.

Using the transfer prior \eqref{eq:direct_transfer}, the total estimation variance equals (both total variance and bias are derived in the appendix \ref{sec:variance}):
\begin{align}\label{eqn:tv}
		\text{total variance} & = \text{tr}(\text{Var}(\widetilde{\nabla f(\mathbf{x})}) \nonumber \\
		& = \frac{1}{B}(\lambda^2(n+1)||\nabla f(\mathbf{x})||_2^2 + 4(1-\lambda)\lambda (\nabla f(\mathbf{x})^T \boldsymbol{e})^2) \nonumber
\end{align}


\begin{equation*}
    \text{Specifically, total variance} = \left\{\begin{array}{ll}
        \frac{n+1}{B}||\nabla f(\mathbf{x})||_2^2, & \text{when } \lambda=1\\
        0, & \text{when } \lambda=0
        \end{array}
        \right.
\end{equation*}

Here, $(n+1)||\nabla f(\mathbf{x})||_2^2 \gg 4 (\nabla f(\mathbf{x})^T \boldsymbol{e})^2 $ as $n$ is typically larger than 1 for high dimensional data such as images and $||\nabla f(\mathbf{x})||_2^2 \geq (\nabla f(\mathbf{x})^T \boldsymbol{e})^2$. \revisionPQ{}{Therefore, $\frac{d \text{tr}(\text{Var}(\widetilde{\nabla f(\mathbf{x})})}{d \lambda} = \frac{2}{B}(((n+1)||\nabla f(\mathbf{x})||_2^2-4(\nabla f(\mathbf{x})^T \boldsymbol{e})^2)\lambda+2(\nabla f(\mathbf{x})^T \boldsymbol{e})^2) \geq 0$ for $\forall \lambda \in [0, 1]$, which means the total variance will monotonically decrease as $\lambda$ decreases. Thus, when $\lambda$ is closer to 0, i.e., the search direction relies more on the prior explanation, the variance will be much lower than the case where no search direction is provided (i.e., setting $\lambda$ close to 1).}

Besides, the estimation $\text{bias} = (1-\lambda)^2 || \nabla f(\mathbf{x})||_2^2(1-\text{cos}(\theta)^2)$, where $\text{cos}(\theta)$ is the cosine similarity between $\boldsymbol{e}$ and $\nabla f(\mathbf{x})$. The bias will decrease as $\text{cos}(\theta)$ increases, which indicates that EGTA is useful for $\boldsymbol e$ that is similar to the gradient, i.e., has a larger cosine similarity to the gradient.

\noindent{\textbf{Explanation-guided Sampling Attack (EGSA):}} Explanations using the absolute value of the gradients may lead to cleaner visualization~\cite{simonyan2013deep, smilkov2017smoothgrad} and the explanation mechanism may return $\mathbf{|e|}$ to the users. In this case, directly using EGTA will lead to a highly biased gradient estimation. To address this we propose EGSA based on the intuition that for locations that has a small gradient value, the corresponding explanation value should also be small. Therefore, we dynamically scale the value of the search vector $\mathbf{u}$ according to the explanation value. Moreover, in appendix \ref{sec:opt_approx}, we theoretically quantify the advantage of EGSA by proving the optimality of the covariance matrix approximation step. 

EGSA can be implemented by replacing \eqref{eq:direct_transfer} with the following\footnote{$\odot$ denotes the Hadamard product.} in Fig. \ref{fig:egta}:

\begin{equation}\label{eq:transfer_sample}
    \mathbf{u}_b = \sqrt{\lambda} \mathbf{u} + \sqrt{1-\lambda} \mathbf{|e|} \odot \mathbf{u}
\end{equation}

%
%
%

\begin{figure}[h]
\includegraphics[width=0.9\linewidth]{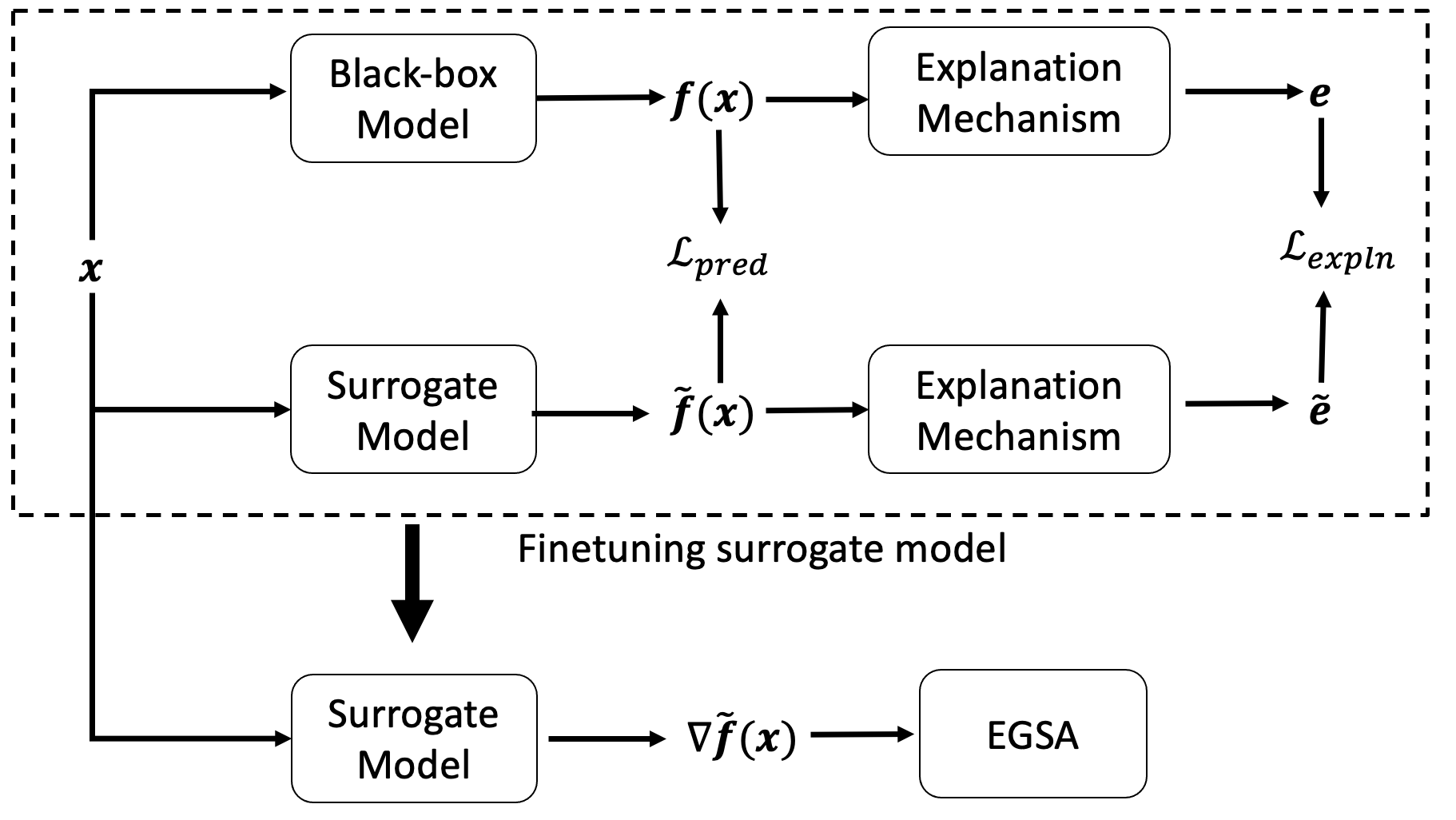}  \centering
	\caption{Explanation-Guided  Surrogate  Model  Attack  (EGSMA)}
	\label{fig:egsma}
	\vspace{-1em}
\end{figure}

\noindent{\textbf{Explanation-Guided Surrogate Model Attack (EGSMA)}} Given an input $\mathbf x$, the black-box model $f(\mathbf x)$, and  explanation $E$, we fine tune a surrogate model $\tilde{f}(\mathbf x)$ for efficient attacks as shown in Fig. \ref{fig:egsma}. In this attack, we assume that the adversary knows the explanation mechanism that is being used by the victim model to generate the explanation. 

We train the surrogate model such that for a given input its prediction matches that of the victim model. To perform training, we use the information available from both the victim model as well as the explanation as follows.

Let $t$ be the label predicted by the victim model. We define the prediction loss of the surrogate model $\mathcal{L}_{pred}\overset{def}{:=}\tilde{f}_t(\mathbf x) - \underset{i\neq t}{\mathrm{max}}\text{ } \tilde{f}_i(\mathbf x)$. Thus, the surrogate model is penalized only when the logit score of the predicted class $t$ denoted by $\tilde{f}_t$, is not the largest among all classes. This will cause the surrogate model to mimic the prediction of the victim model. In addition, we also force the explanation provided by the surrogate model to be close to the true explanation by minimizing $\mathcal{L}_{expln}\overset{def}{:=}||{\mathbf e}-\mathbf{\tilde{e}}||$. Here ${\mathbf e}$ is produced by the victim model and $\tilde{\mathbf e}$ is generated using the gradient vector from the surrogate model. 

We combine the prediction and explanation losses to train the surrogate model using $\mathcal{L}_{sur}$ loss defined as:
\begin{equation}\label{eqn:egsma_loss}
    \mathcal{L}_{sur} = \mathcal{L}_{pred} + \alpha \mathcal{L}_{expln}
\end{equation}

After fine tuning the surrogate model on the loss Eqn.~\eqref{eqn:egsma_loss} with $\alpha \in [0, 1]$, we use EGSA to find the adversarial example.

\begin{figure}[h]
\includegraphics[width=0.98\linewidth]{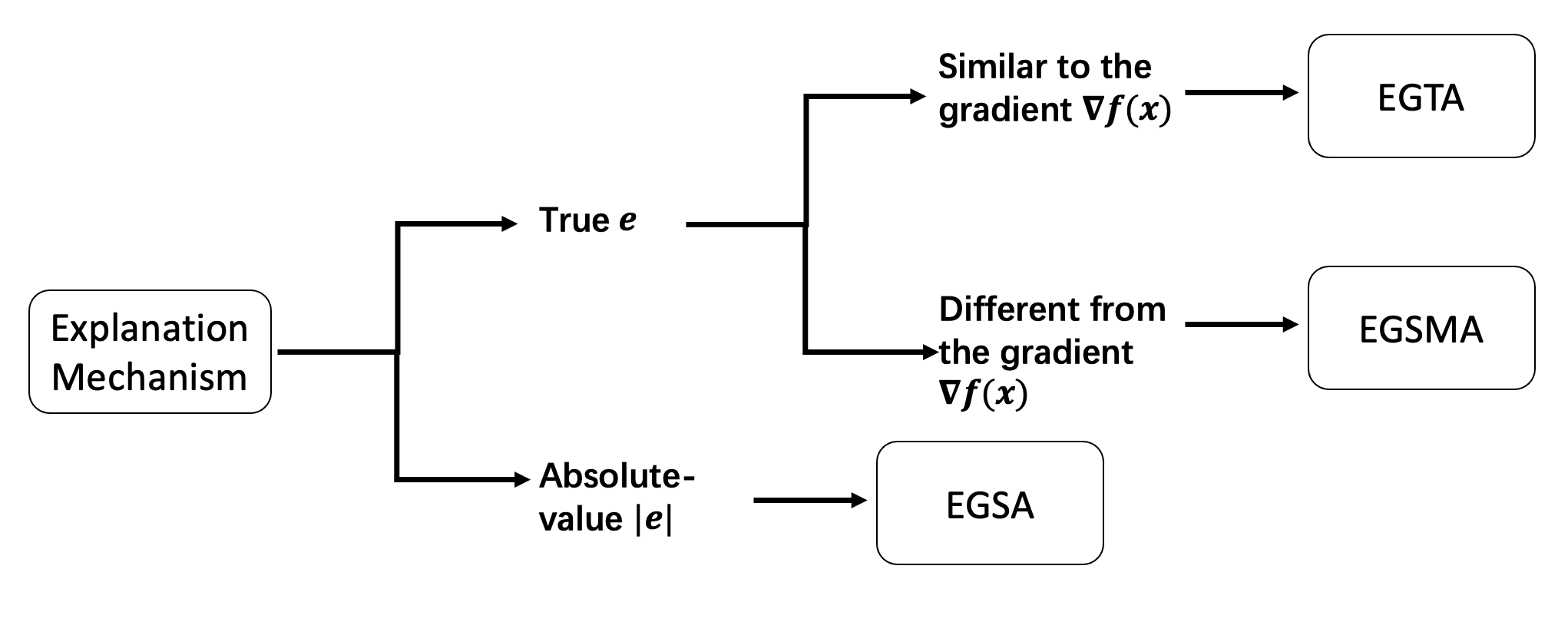}  \centering
	\caption{Choosing the attack strategy based on explanation type}
	\label{fig:conditions}
\end{figure}

\noindent{{\bf Choosing the best attack strategy:}}
\revisionPQ{}{We provide a taxonomy of the attacks identifying the conditions under which they perform the best in Fig.\ref{fig:conditions}. For distinction, we call $\boldsymbol{e}$, which is directly produced from explanation mechanism, as a \textit{true explanation}, and $|\boldsymbol{e}|$ as the \textit{absolute-value explanation} as shown in Table \ref{table:notations}}. (\romannumeral 1) If the true explanation is available, the attacker can use EGTA when the explanation is similar to the gradient; (\romannumeral 2) If the true explanation is available but the form is different from the true gradient, the attacker can use EGSMA. (\romannumeral 3) If only the absolute-value explanation is accessible, the attacker can use EGSA. 
By inspecting the returned explanations, the attacker can know which form of the explanation is used so he can choose the attack strategy based on the explanation type.

\section{Membership Inference Attack (MIA)}
 In this attack, the goal of an adversary is to determine if a given data sample was used to train an ML model. Successful membership inference attacks can compromise user's privacy. For instance, if a machine learning model is trained on the data collected from people with certain characteristics, by knowing that a victim’s data sample belongs to the training data of the model, the attacker can immediately learn sensitive information about the victim~\cite{shokri2017membership, sablayrolles2019white}.

\noindent {\bf Notation:} Let ${\bf x} \in \mathbb{R}^n$ be the input, and $F(\bf x)$ denote the victim model. $\mathcal{D}_{train}$ is the dataset used to train $F(\bf x)$. $\mathcal{D}_{out}$ contains the set of $\bf x$ that are not in $\mathcal{D}_{train}$.

\noindent {\bf Adversary Access:} The attacker can provide $\bf x$ to the black-box model and obtain the prediction $F(\bf x)$ and its explanation $E(\bf x)$ in a black-box setting. Table \ref{table:requirements} summarizes the additional requirements to perform membership inference attacks for existing attack methods \cite{sablayrolles2019white, yeom2018privacy, choquette2021label, shokri2021privacy}.

\noindent {\bf Attack Objective:} Given a sample $\bf x$, the attacker aims to infer whether the sample belongs to the dataset $\mathcal{D}_{train}$ or $\mathcal{D}_{out}$.

\subsection{Our Attack: Revisiting MIA Using Explanations}
For data points that are in the training set, the victim model is certain about the prediction as the training algorithms are designed to push the points away from the decision boundary. In a white-box setting, this is indicated by a relatively low loss score on training samples, which is used to determine membership in the training data. Consequently, the gradient values computed using the loss should also be fairly small for the training samples. This intuition is used in \textit{OPT-var}~\cite{shokri2021privacy}, to mount MIA in black-box settings when explanations are available. They show that variance of the explanation $\boldsymbol{E}(F, \bf x)$ produced from the logit score $F(\bf x)$ can indicate membership.


We build on the intuition in \textit{OPT-var}~\cite{shokri2021privacy} 
but note that the logit score does not completely capture the victim model's prediction confidence.
The reason is that the logit score of a particular class does not take into account the logit scores of other classes. 
\revisionPQ{}{On the other hand, the softmax function of the logit consider the interaction of classes and the corresponding gradient may contain richer membership information.} Therefore, we propose to use the variance of explanation produced from the softmax score $f(\bf x) = \boldsymbol{\sigma}(F(\bf x))$ as an indicator of membership, where $\boldsymbol{\sigma}(F(\bf x))$ is calculated as follows:
\begin{equation}
    \boldsymbol{\sigma}(F(\bf x))_i = \frac{e^{F(\bf x)_i}}{\sum_{j=1}^k{e^{F(\bf x)_j}}}
\end{equation}

The variance of explanation $\boldsymbol e=\boldsymbol{E}(\boldsymbol f, \bf x)$ is calculated as:
\begin{equation}
    \text{Var}(\boldsymbol e) = \sum_{i=1}^n (\boldsymbol{e}_i - \frac{1}{n}\sum_{j=1}^n\boldsymbol{e}_j)
\end{equation}

where $\bf{x}$ and $ \bf{e} \in \mathbb{R}^n$.
\section{Model Extraction attack (MEA)}

Despite the fact that machine learning and neural networks are widely applied in industry settings, the trained models are costly to obtain. Furthermore, there are security \cite{carlini2017towards, carlini2017adversarial} and privacy \cite{shokri2017membership, fredrikson2015model} concerns for revealing trained models to potential adversaries. Thus, trained model should be treated as proprietary and protected the same way. Unfortunately, prior works \cite{tramer2016stealing, orekondy2019knockoff, milli2019model} have shown that an adversary with query access to a model can steal the model to obtain a copy that is \revisionPQ{}{\textit{functionally equivalent}, i.e., predictions are identical to those of the victim model on all possible inputs.} In particular, \cite{milli2019model} propose gradient matching to improve MEA. However, there is no systematic study regarding how various explanations \cite{simonyan2013deep, sundararajan2017axiomatic, ribeiro2016should} can help mount MEA.

\noindent{\bf Notation:} Let ${\bf x} \in \mathbb{R}^n$ be the input, $F(\boldsymbol x)$ denote the victim model and $\hat{F}(\boldsymbol x)$ represent the extracted model.

\noindent {\bf Adversary Access:} Model extraction attacks target the confidentiality of a victim model deployed on a remote server. We assume the adversary knows the architecture of the victim model but the corresponding parameter values are unknown. We also assume that the attacker possess a subset of the training data $\mathcal{S}_{train}$. However, the adversary does not necessarily have the ground-truth labels and can query $F(\boldsymbol x)$ for obtaining the labels. Through black-box query access to the victim model $F(\boldsymbol x)$ and the corresponding explanation $E(\boldsymbol x)$, the attacker aims at producing a \textit{functionally equivalent} model: the extracted model achieves same performance on the same test dataset $\mathcal{S}_{test}$, i.e., $\forall \boldsymbol x \in \mathcal{S}_{test}$, $\hat{F}(\boldsymbol x) = F(\boldsymbol x)$.

\subsection{Our Attack: Explanation-matching Strategy}
Here we follow the strategy proposed in~\cite{milli2019model} to match the predictions and the explanations produced by the victim model $F(\boldsymbol x)$. We denote the extracted model by the adversary as $\hat{F}(\boldsymbol x)$ and the corresponding explanation as $E(\boldsymbol x)$. The adversary can perform MEA by minimizing the prediction loss, $\mathcal{L}_{P} = ||\hat{F}(\boldsymbol x) - F(\boldsymbol x)||$, and an explanation matching loss, $\mathcal{L}_{E} = ||\hat{E}(\boldsymbol x) - E(\boldsymbol x)||$. In our experiments, we minimized $\mathcal{L}_{J} = \mathcal{L}_{P} +  \alpha \mathcal{L}_{E}$ where $\alpha \in [0, 1]$. For LIME, we perform prediction matching with the interpretable model.
\section{Experiments}\label{sec:exps}
\subsection{Evasion attack (EA)}
We use the number of queries needed to achieve a particular attack success rate as the metric to measure attack efficiency across all experiments for both soft and hard-label attacks. The baseline attack is when the adversary has access to only the model output and no explanation. All our experiments have been performed on a ResNet-50~\cite{he2016deep} as the target model trained on ImageNet~\cite{deng2009imagenet} dataset. The reported values are obtained by averaging over successful attacks on $100$ correctly classified images. For the $\lambda$ used in \eqref{eq:direct_transfer}, we fix its value as 0.9 in the remaining experiments. 

In summary, we consider the true explanation and the absolute-value explanations returned to the attacker. The attacker will use the three proposed explanation-guided attack approaches. We omit the absolute-value explanations for EGTA and EGSMA due to inaccurate transfer information. We also exclude the true explanation for EGSA because it has the same performance as the absolute-value explanation. For EGSMA, We omit the results for LIME since the loss of $\mathcal{L}_{expln}$ is not differentiable over the surrogate model parameters and hence EGSMA cannot be directly applied.



\begin{figure*}[t]
    \centering
    \begin{minipage}{0.3\linewidth}
     \centering

    \includegraphics[width=\linewidth]{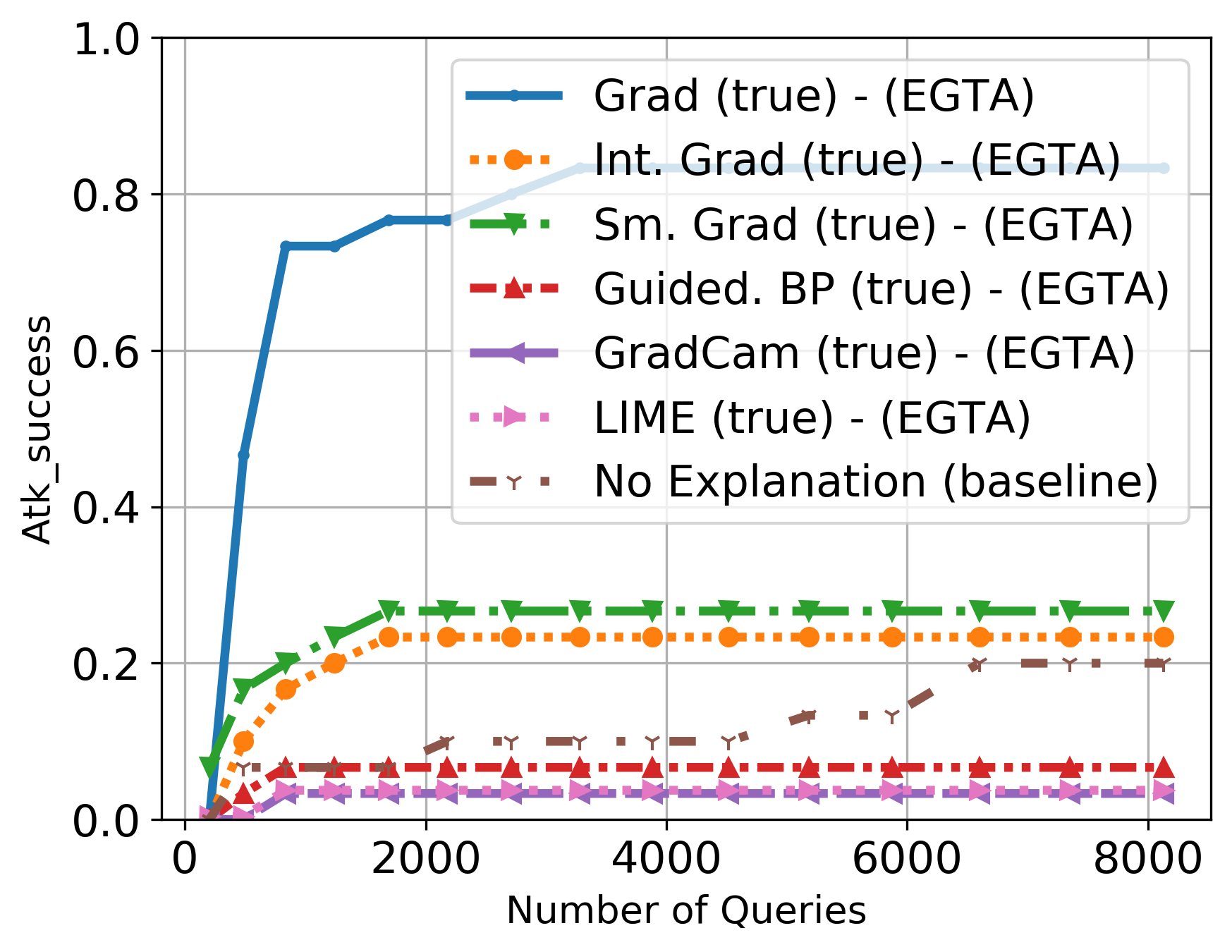} \centering (a) Soft-label\par
    \includegraphics[width=\linewidth]{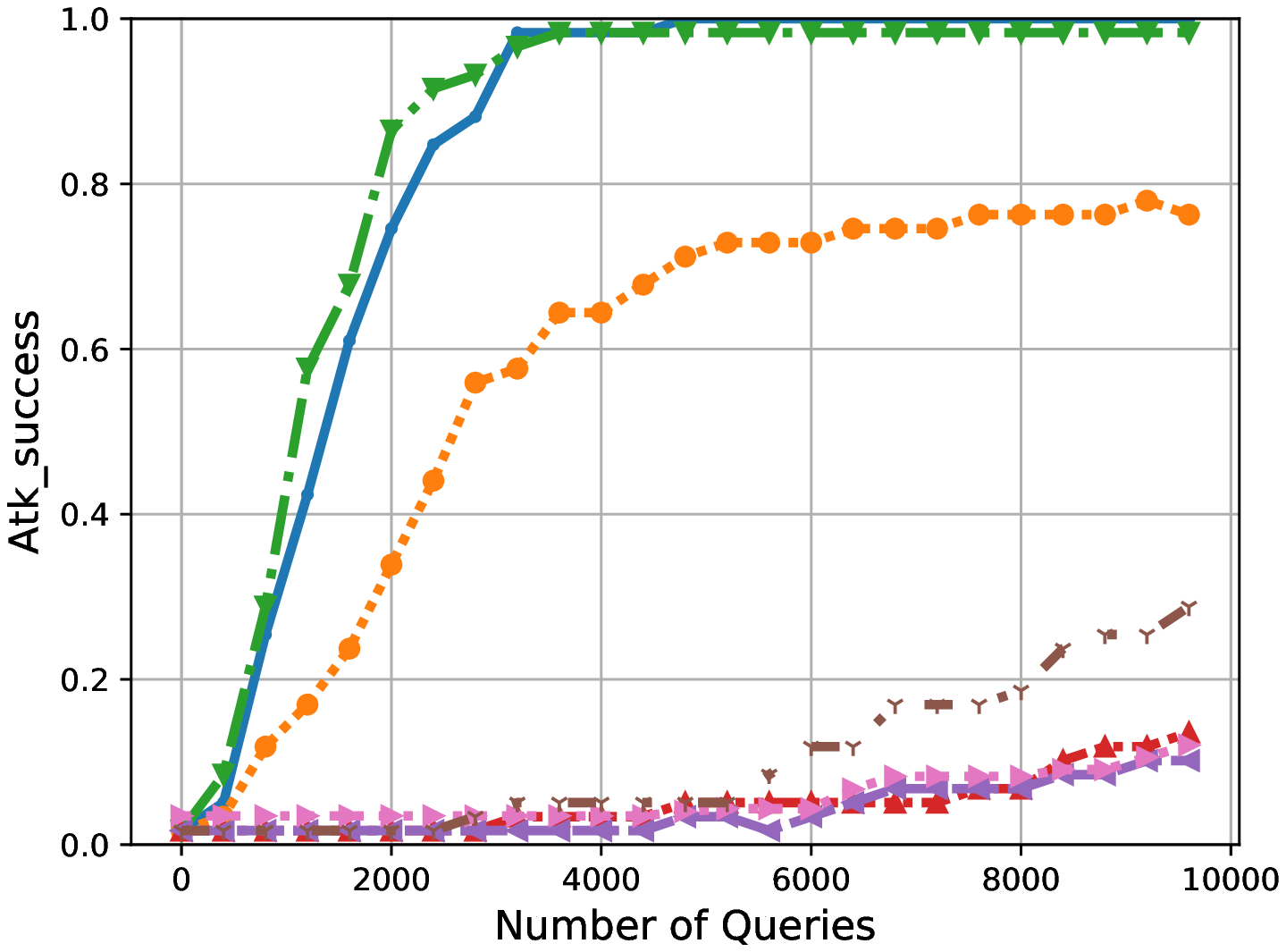} \centering (b) hard-label\par
    
    \caption{Attack success rate of EGTA using the true explanations.}
    \label{fig:exps-soft-org}
  \end{minipage}
    \begin{minipage}{0.3\linewidth}
     \centering
  \includegraphics[width=\linewidth]{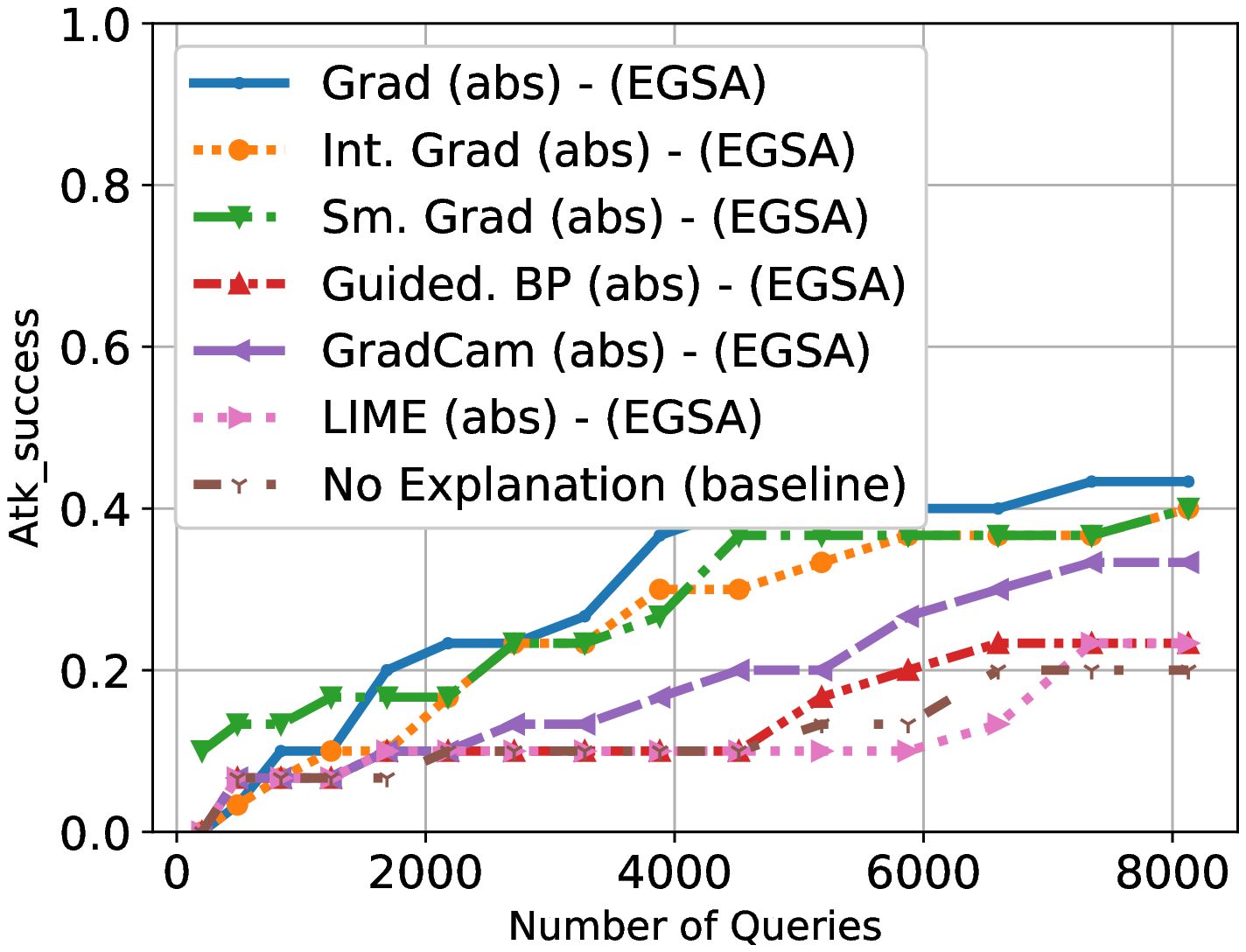} \centering (a) Soft-label \par
  \includegraphics[width=\linewidth]{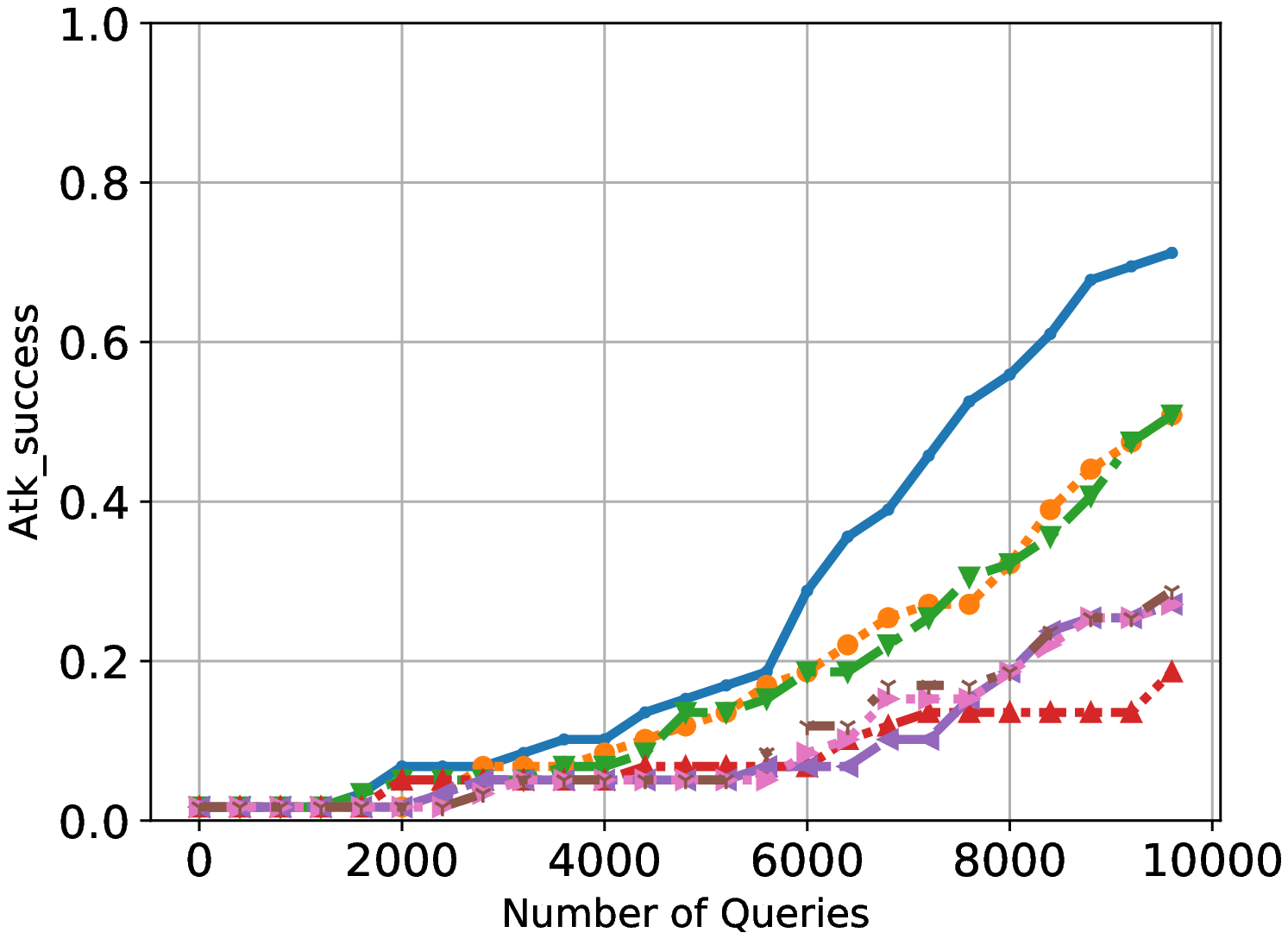} \centering (b) Hard-label\par
    \caption{Attack success rate of EGSA using explanations with absolute values.}
    \label{fig:exps-soft-abs}
  \end{minipage}
    \begin{minipage}{0.3\linewidth}
     \centering
     \includegraphics[width=\linewidth]{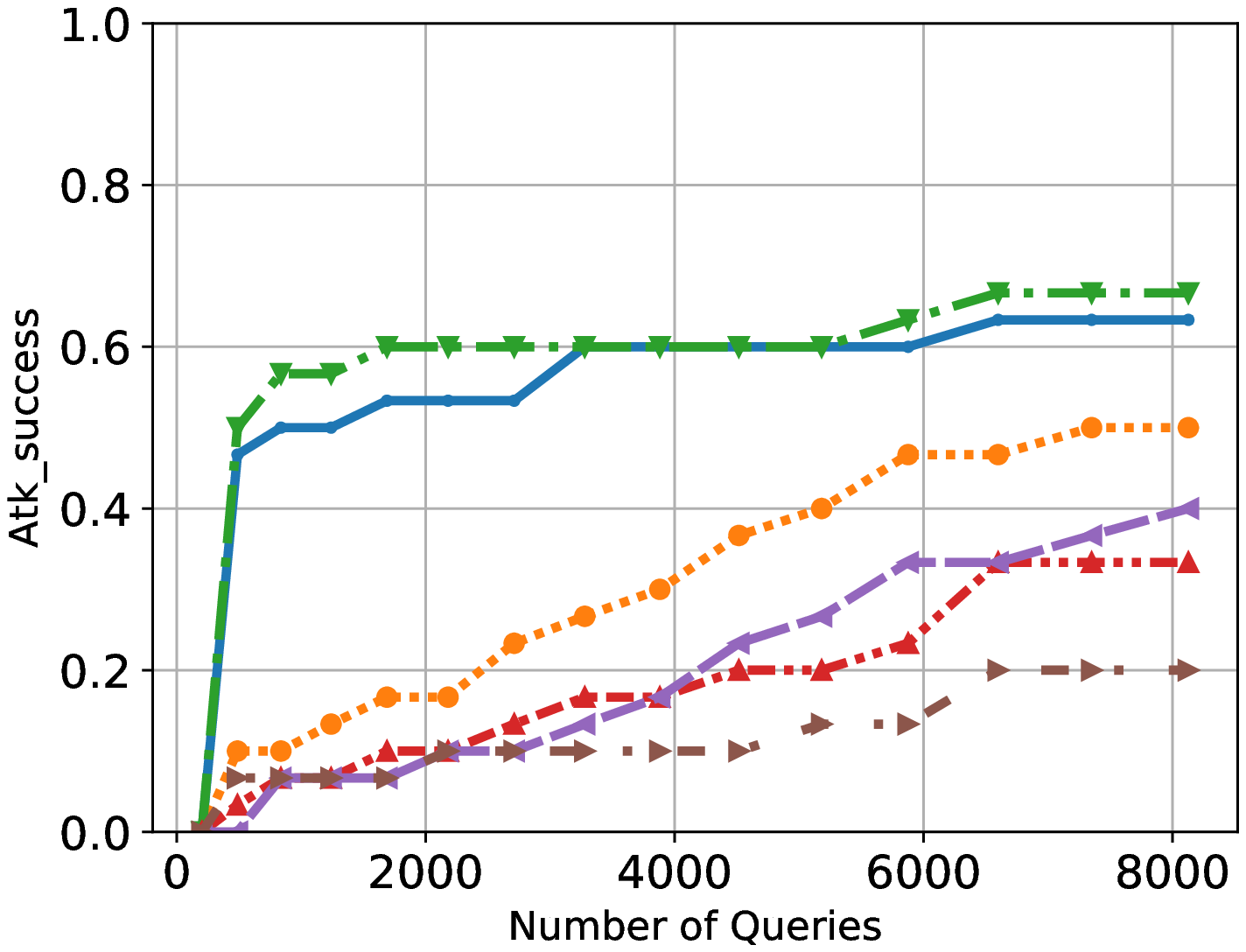} \centering (a) Soft-label \par
     \includegraphics[width=\linewidth]{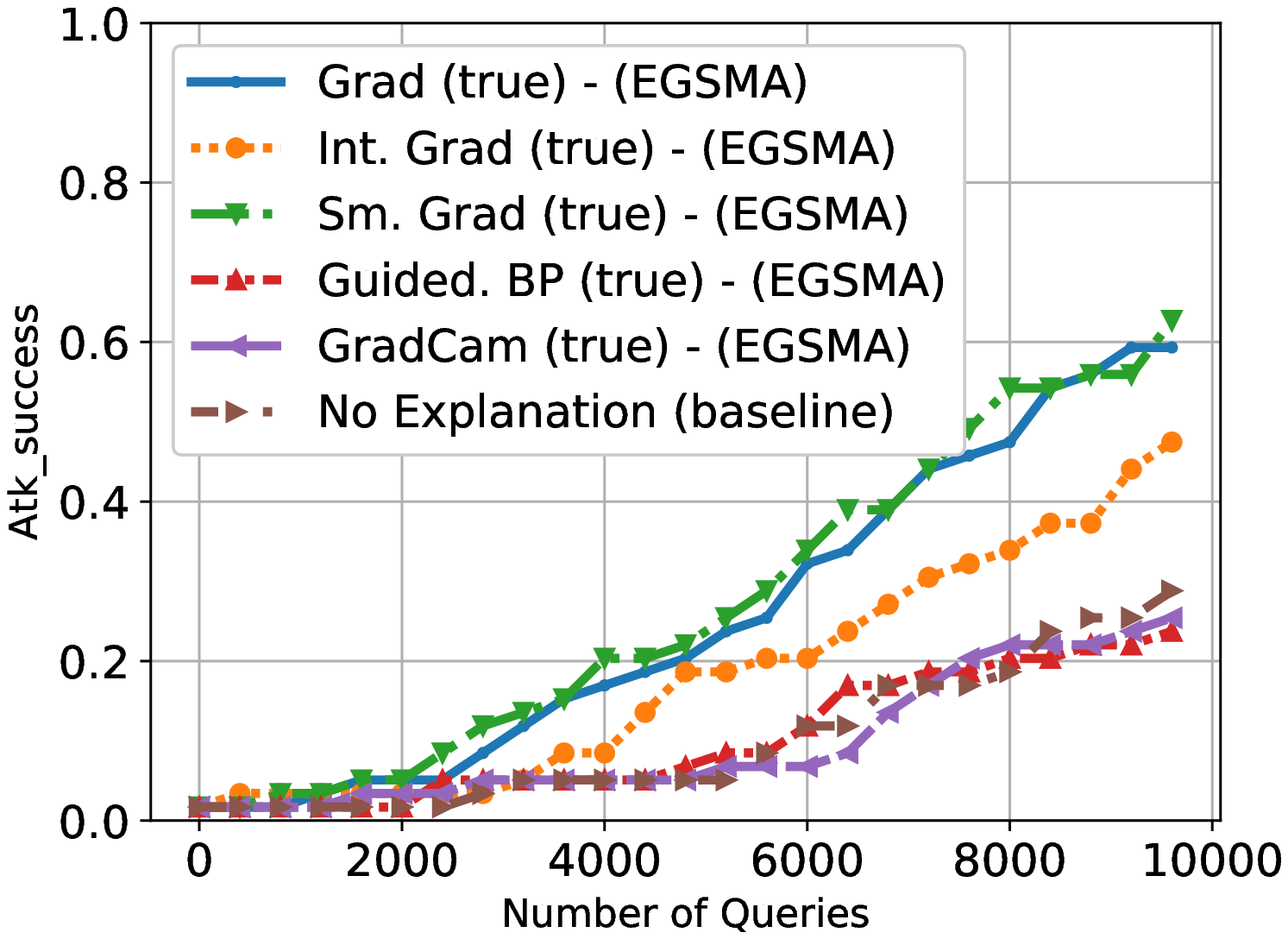} \centering (b) Hard-label\par
    \caption{Attack success rate of EGSMA using true explanations.}
    \label{fig:exps-learn-s}
  \end{minipage}
\end{figure*}

\subsubsection{Explanation-guided Transfer Attack (EGTA)}

\noindent \textbf{Soft-label attack results using true explanations:} We start with an image in the target class $t$ and try to generate perturbation such that it is misclassified to any other class by the model. We set $L_2=1.25$ as our perturbation budget to compare with \cite{cheng2019improving}. In Fig. \ref{fig:exps-soft-org} (a), we show the median attack success rate of achieving the target distortion with various queries consumed. First, we can see that our proposed approach of using the information from the Gradient, Integrated Gradient, and SmoothGrad can reduce the query budget by more than $\times 10$, indicating that gradient information is leaked by the above explanations. Besides, since this is an biased estimator, using Integrated Gradient and SmoothGrad explanations leads to suboptimal convergence. Hence, the attack success rate quickly saturates due to the bias in gradient estimation. Also, for explanations drastically different from Gradient, such as Guided Backprop and GradCam, the direct transfer attack (EGTA) may lead to poor attack performance due to biased gradient estimation.



\noindent \textbf{Hard-label attack results using true explanations:} In this setting, we consider a targeted attack. An attacker starts with an image in the target class and iteratively reduces the $L_2$ distance to the original image $\mathbf{x}$ to generate the adversarial image. Here we use the distortion threshold $L_2=12.5$ as the attack success metric because we empirically found that the noise magnitude is imperceptible. Fig. \ref{fig:exps-soft-org} (b) shows that having access to the explanation reduces the number of queries needed to reach the same attack success rate. For Gradient, IntegratedGradients, and Smooth Grad, the transfer attack can reduce the query budget by about 10 times. However, for other variants such as Guided Backprop and GradCam, directly using transfer attack leads to worse attack performance.


\subsubsection{Explanation-Guided Sampling Attack  (EGSA)}
\noindent \textbf{Soft-label attack results using absolute-valued explanation:} The previous EGTA is only applicable to true explanations, but does not generalize to explanations with absolute values for visualization. Here we demonstrate that EGSA applies to the absolute-value explanations in Fig. \ref{fig:exps-soft-abs} (a). We can observe that the attack efficiency is significantly improved for Gradient, Smooth Grad, and Integrated Gradients by 3 -5 times. The improvement is limited for LIME, Guided Backprop, and GradCam. 

\noindent \textbf{Hard-label attack results using absolute-valued explanation:} Fig. \ref{fig:exps-soft-abs} (b) shows the attack efficiency by leveraging absolute-valued explanations. Similarly, we can observe that the attack efficiency is significantly improved for Gradient, Smooth Grad, and Integrated Gradients by 2 times. The improvement is limited for LIME, Guided Backprop, and GradCam. 


\subsubsection{Explanation-Guided  Surrogate  Model  Attack  (EGSMA)} 
\noindent \textbf{Soft-label attack results using true explanations:} In Fig \ref{fig:exps-learn-s} (a), we show the performance of EGSMA using various explanations. First, we can observe that EGSMA can robustly increase the attack performance by 10 times for Smooth Grad, and by 2-3 times for Integrated Gradients, GradCam, and Guided Backprop. Here, EGSMA can increase the performance of attacks when the direct transfer fails. Fig. \ref{fig:exps-learn-s} (a) also shows that EGSMA is a robust method where the attacks success rate increase consistently compared to EGTA where the direct transfer fails.

\noindent \textbf{Hard-label attack results using true explanations:} Fig \ref{fig:exps-learn-s} (b) shows the performance of EGSMA in hard-label setting. It produces 2 - 3 times improvment for SmoothGrad and IntegratedGradients.


\subsubsection{Attack results summary}

\begin{table*}[h]
	\centering
	\begin{adjustbox}{max width=0.85\textwidth}
		\begin{tabular}{c|c|cccc|cccc|cccc}
			\toprule
			Atk. rate  & {None} & \multicolumn{4}{c}{Smooth Grad } & \multicolumn{4}{c}{Grad} & \multicolumn{4}{c}{Int. Grad} \\
			\midrule
			 Atk. strategy & - & $\text{P-RGF}_D$ \cite{cheng2019improving} & EGTA & EGSA & EGSMA & $\text{P-RGF}_D$ \cite{cheng2019improving} & EGTA & EGSA & EGSMA & $\text{P-RGF}_D$ \cite{cheng2019improving} & EGTA & EGSA & EGSMA \\
			\midrule
			1k & 6.7 & 16.7& 26.7 & 17.7 & \bf 60.0 & 50.0 & \bf 76.7 &  10.0 & 53.3 & 20.0 &  \bf 23.3 & 6.7 & 11.7 \\
			\midrule
            4k & 10 & 23.3 & 26.7 & 27.1 & \bf 60.0 & 56.7 &  \bf 76.7 & 23.3 & 60.0 & \bf 23.3 & \bf 23.3 & 16.7 & 17.7 \\
            \midrule
            8k & 20 & 33.3 & 26.7 & 48.6 & \bf 66.7 & 67.7 & \bf 83.3 & 46.7 & 63.3 & 33.3 &  23.3 & 33.3 & \bf 53.3 \\
			\midrule
			\bottomrule
		\end{tabular}
	\end{adjustbox}
	\caption{Performance comparison on the soft-label attack success rate. Note that EGSA will give the same results for the true and absolute-valued explanations. $\text{P-RGF}_D$ \cite{cheng2019improving} uses various explanation as the prior, e.g, SmoothGrad. Notice that simply relies on the the transfer prior \cite{cheng2019improving} may lead to suboptimal attack due to the bias estimation of gradients.}
	\label{table:attack_succ_rate_s}
	\vspace{-1em}
\end{table*}

\begin{table*}[h]
	\centering
	\begin{adjustbox}{max width=0.8\textwidth}
		\begin{tabular}{c|c|c|ccc|ccc|ccc}
			\toprule
			Atk. rate  & {None} & QEBA \cite{li2020qeba} & \multicolumn{3}{c}{Smooth Grad} & \multicolumn{3}{c}{Grad} & \multicolumn{3}{c}{Int. Grad} \\
			\midrule
			 Atk. strategy & - & - & EGTA & EGSA & EGSMA & EGTA & EGSA & EGSMA & EGTA & EGSA & EGSMA \\
             \midrule
             1k & 1.7 & 1.7 & \bf 68.8 & 1.7 & 5.1 & \bf 61.0 & 1.7 & 6.3 & \bf 23.8 & 1.7 & 3.4 \\
             \midrule
             4k & 5.1 & 6.8 & \bf 96.6 & 5.1 & 13.5 & \bf 98.3 & 8.5 & 12.0 & \bf 57.6 & 6.8 & 5.1 \\
			\midrule
            8k & 18.6 & 39.0 & \bf 98.3 & 32.2 & 54.2 & \bf 100 & 55.9 & 46.6 & \bf 76.3 & 32.2 & 33.9 \\
            \midrule
			\bottomrule
		\end{tabular}
	\end{adjustbox}
	\caption{Performance comparison on the hard-label attack success rate. QEBA \cite{li2020qeba} utilizes the gradient from 5 different surrogate models, such as ResNet-50, DenseNet-121, etc.}
	\label{table:attack_succ_rate_h}
	\vspace{-1em}
\end{table*}

Here we provide the following major guidelines to select different methods for different available explanations to perform EA based on the results:
\begin{itemize}
    \item For explanations that are close to the gradient, such as Smooth Grad and IntegratedGradients, it is recommended to apply EGTA to produce adversarial examples.
    \item For true explanations that are variants of the Gradient, including Guided Backprop and GradCam, the surrogate-model based attack EGSMA performs better. Besides, Table \ref{table:attack_succ_rate_s} and \ref{table:attack_succ_rate_h} show that EGSMA outperform the previous state-of-the-art transfer prior attack $\text{P-RGF}_D$ \cite{cheng2019improving} and QEBA \cite{li2020qeba}.
    \item For absolute-value explanations, the sampling-based attack EGSA can be applied to improve attack effectiveness. 
\end{itemize}

\subsection{Membership Inference Attack}
Here we construct the membership inference attack by measuring the accuracy of the membership inferenced by the attackers \cite{shokri2017membership, shokri2021privacy, yeom2018privacy} on 5 datasets: \textbf{CIFAR-10\cite{krizhevsky2009learning}}, \textbf{Purchases\cite{shokri2017membership}}, \textbf{Texas. \cite{shokri2017membership}}, \textbf{HAR\cite{li2021units, roggen2010collecting}}, and \textbf{Covid \cite{rahman_2021}}. Details about the datasets are provided in the appendix \ref{sec:dataset}.

\begin{table}[]
\begin{adjustbox}{max width=0.45\textwidth}
    \begin{tabular}{c|c|c|c|c}
    \toprule
    Works             & Ground-truth label & Probability prediction & Excessive queries & Explanation \\ \midrule
    \textit{MALT} \cite{sablayrolles2019white} &  \cmark      &      \cmark        &    \xmark       &   \xmark   \\ \midrule
    \textit{GAP-attack} \cite{yeom2018privacy}        &     \cmark      &      \xmark        &    \xmark       &   \xmark    \\ \midrule
    \textit{Boundary-attack} \cite{choquette2021label}  &     \cmark      &      \xmark        &    \cmark       &   \xmark       \\ \midrule
    \textit{OPT-var} \cite{shokri2021privacy}     &     \xmark      &      \xmark        &    \xmark       &   \cmark      \\ \midrule
    Ours       &     \xmark      &      \xmark        &    \xmark       &   \cmark      \\ \bottomrule
    \end{tabular}
\end{adjustbox}
\caption{The requirements for performing MIA}\label{table:requirements}
\vspace{-2em}
\end{table}

\begin{table*}[h]
	\centering
	\begin{adjustbox}{max width=0.9\textwidth}
		\begin{tabular}{c|c|c|c|c|c|c|c|c|c|c}
			\toprule
			Accuracy & \textit{MALT} \cite{sablayrolles2019white} & \textit{GAP-attack} \cite{yeom2018privacy} & \textit{Boundary-attack} \cite{choquette2021label} & \textit{OPT-var}\cite{shokri2021privacy} & OPT-Grad. & OPT-Int. Grad & OPT-Smooth Grad  & OPT-Guided Backprop & OPT-GradCam & OPT-LIME\\
            \midrule
            Covid & 69.0 & 55.7 & 64.0 & 51.8 & 66.8 & \bf 67.2 & 66.8 & 67.2 & 66.7 & 58.5 \\
            \midrule
            CIFAR-10 & 76.75 & 55.9 & 71.1 & 51 & 70 & 71.3 & 71.0 & 70.0 & 71.5 & \bf 73.8\\
            \midrule
            HAR & 60.0 & 56.2 & 56.8 & 48.9 & \bf 58.9 & 58.2 & 58.8 & 58.8 & 58.5 & - \\
            \midrule
            Purchase & 65.3 & 55.7 & 55.9 & 56.2 & 63.4 & \bf 63.7 & 63.5 & 63.4 & - & 62.3 \\
            \midrule
            Texas & 82.3 & \bf 77.8 & 77.1 & 64.2 & 73.0 & 72.2 & 71.5 & 73.0 & - & 76.0 \\ 
			\midrule
			\bottomrule
		\end{tabular}
	\end{adjustbox}
	\caption{Performance comparison on membership inference attack. Note that ours can achieve better accuracy than \textit{GAP-attack} \cite{yeom2018privacy} and \textit{Boundary-attack} \cite{choquette2021label}, and is comparable to \textit{MALT} \cite{sablayrolles2019white}, without additional attack access except explanations as shown in Table \ref{table:requirements}. Ours also outperform \textit{OPT-var}\cite{shokri2021privacy} using the same attack access.}
	\label{table:membership}
	\vspace{-2em}
\end{table*}

We randomly choose 8000 training samples from the private dataset and 8000 unseen samples for CIFAR-10, Purchase, and Texas. Due to the size of the training set of HAR and Covid, we choose 1000 private training samples and 1000 unseen samples. For each attack algorithm, the adversary also has 200 samples from the private training set to finetune the hyper-parameters of the attack algorithms. Table \ref{table:requirements} shows the required conditions to perform each attack, and Table \ref{table:membership} shows the results for membership inference attack using various attack methods:





\textit{MALT} \cite{sablayrolles2019white} has shown that using loss as the indication of MIA is optimal given an optimal threshold under some assumptions, and it becomes the upper bound of MIA. \textit{GAP-attack} \cite{yeom2018privacy} predicts any misclassified data point as a non-member of the training set and vice versa. This is a naive baseline for label-only membership inference attacks. \textit{GAP-attack} only works well on models that have a relatively lower classification accuracy, e.g., Texas dataset. \textit{Boundary-attack} \cite{choquette2021label} uses the estimated distance to the decision boundary as an indication of membership. The attack works consistently well among all the datasets. However, we also need to point out that \textit{Boundary-attack} requires a harder condition to be performed, as shown in Table \ref{table:requirements}. It needs access to the ground-truth label of a sample of interest and requires thousands of more queries to estimate decision boundary, which may render a costly attack. As is reported in \cite{shokri2021privacy}, \textit{OPT-var} is not a strong indication for Covid, CIFAR-10, and HAR datasets but informative for Purchase and Texas datasets.

After revisiting the relation between explanations and MIA, we found that explanations actually carry much richer membership information instead of what is claimed by \textit{OPT-var} \cite{shokri2021privacy}. Here we define the advantage as $p_{\text{acc}} - p_{\text{rand}} = p_{\text{acc}} - 50$, i.e., the improvement of accuracy over random guess 50\%, and we observe a significant increase compared to the previous \textit{OPT-var}. For instance, OPT-Grad outperforms \textit{OPT-var} consistently by more than 100\%. Besides, Smooth Grad, which is treated as perturbation-based explanation methods and criticized for not following the distribution of the training data \cite{shokri2021privacy}, performs constantly well as other gradient-based explanations. This indicates that Smooth Grad is also authentic to the model, and the membership information is also maintained by such explanations. Furthermore, even region-based explanation Grad-Cam also contains informative membership information. Lastly, in contrast to the claim in \cite{shokri2021privacy} that LIME explanation does not perform better than a random guess, the LIME explanations outperform many other approaches and result in meaningful membership indications. Please note that GradCam cannot be applied naturally to Purchase and Texas because the trained neural networks only consist of a fully connected layer and do not use multiple channels and pooling operations. Besides, we also omitted to apply LIME to time-series data because LIME cannot be naturally extended to this modality.

\subsection{Model Extraction attack (MEA)}
Here we show the results of MEA on 4 different datasets: CIFAR-10, MNIST~\cite{deng2012mnist}, Covid, and HAR. The first threes are on the image domain, and HAR is time-series data. Following \cite{tramer2016stealing}, we plot the test mismatch on the test dataset as the followings:
\begin{equation}
    R_{\text{test}} = T_{\text{test}} - \hat{T}_{\text{test}}
\end{equation}
where $\hat{T}_{\text{test}}$ is the test accuracy of the extracted model $\hat{F}(\boldsymbol x)$ obtained using simple prediction matching or the explanation matching. $T_{\text{test}}$ is the test accuracy of the victim model $F(\boldsymbol x)$ on the test dataset. $R_{\text{test}}$ is the measure of how effective the MEA mechanism is to produce a model with equivalent task accuracy, \revisionPQ{}{which works as a proxy for \textit{functionally equivalent}.}

\begin{figure}[h]
    \minipage{0.24\textwidth}
        \includegraphics[width=\linewidth]{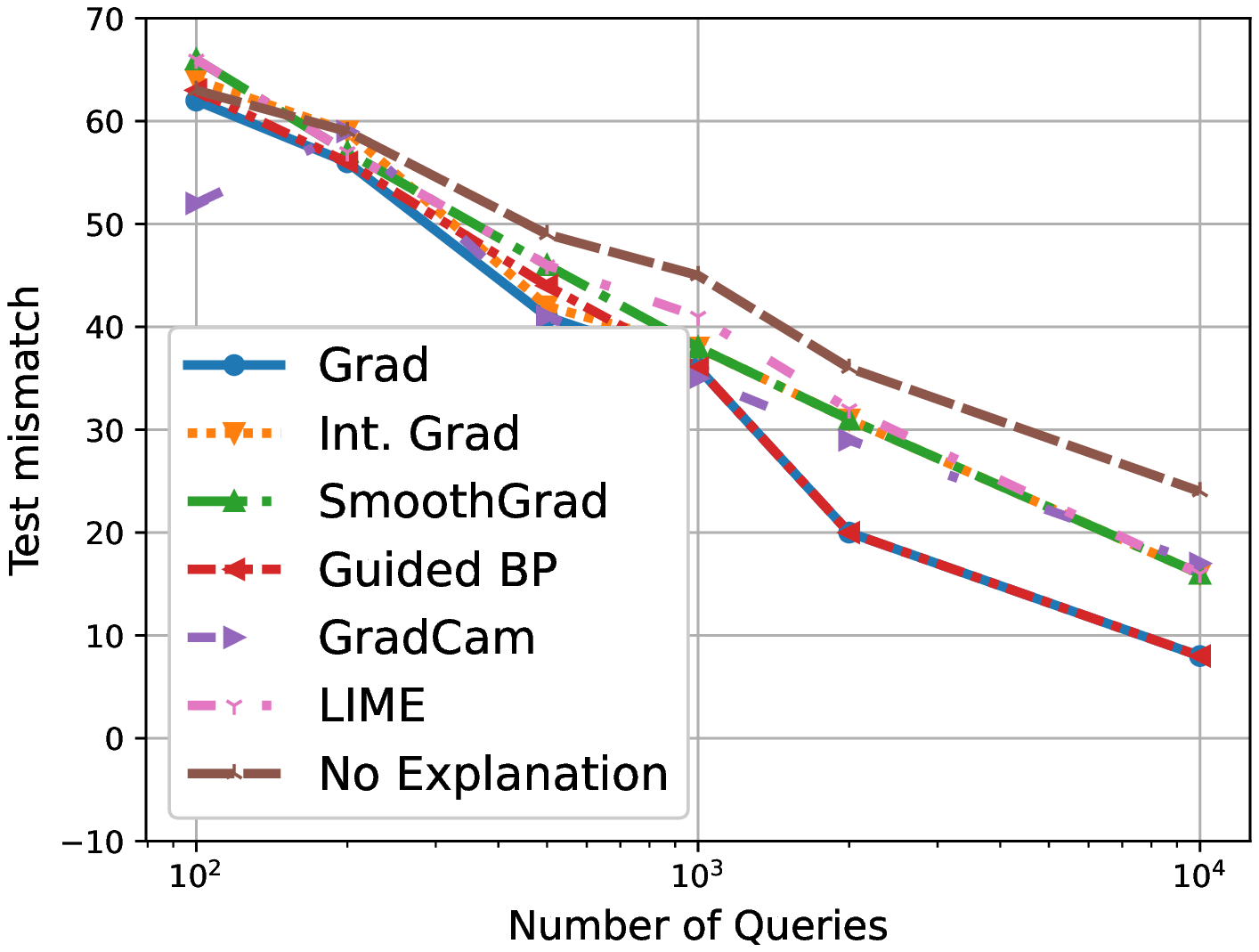} (a) CIFAR-10 \centering \par
    \endminipage\hfill
    \minipage{0.24\textwidth}
        \includegraphics[width=\linewidth]{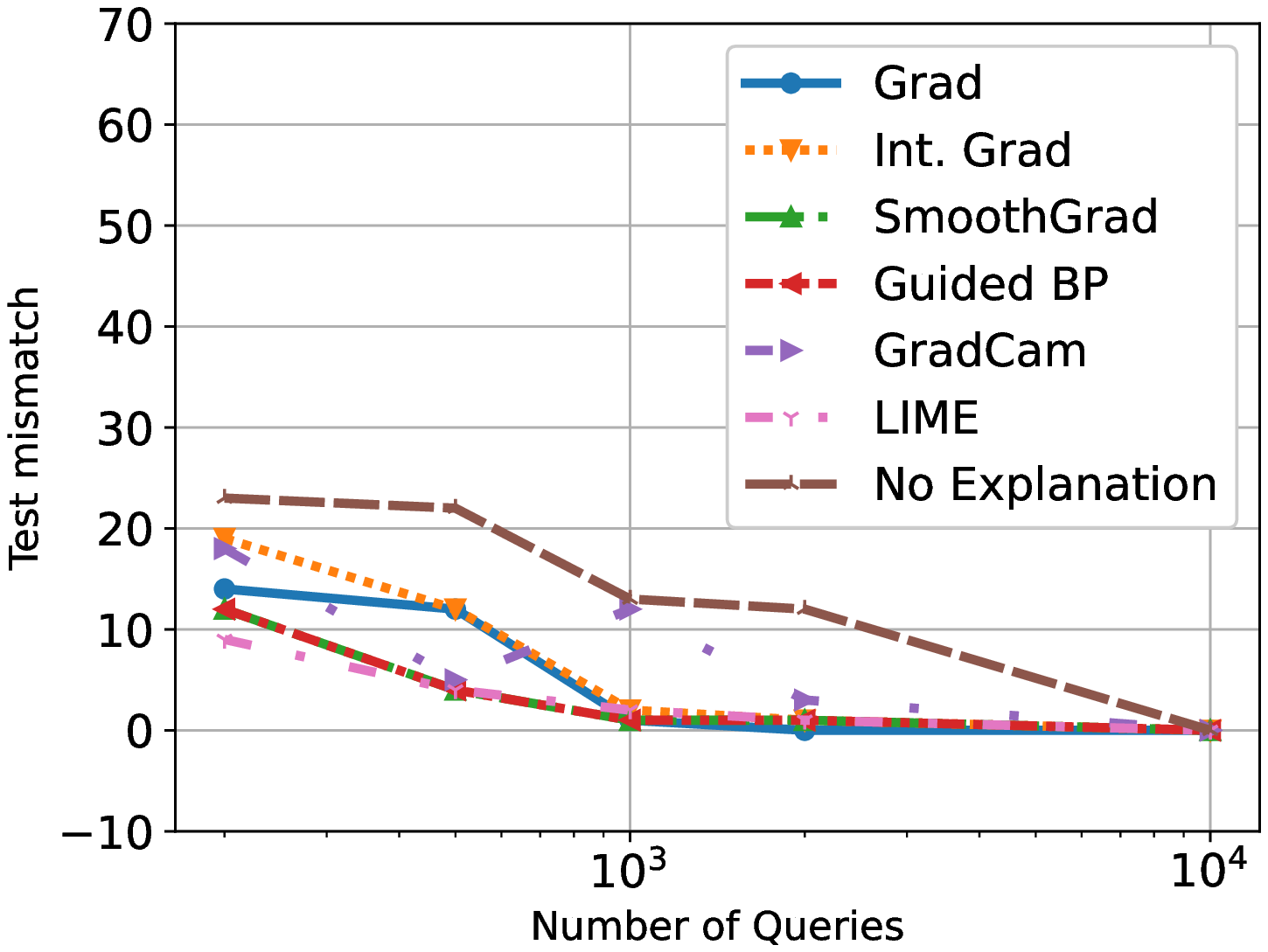}  (b) MNIST \centering \par
    \endminipage\hfill
    
    \minipage{0.24\textwidth}
        \includegraphics[width=\linewidth]{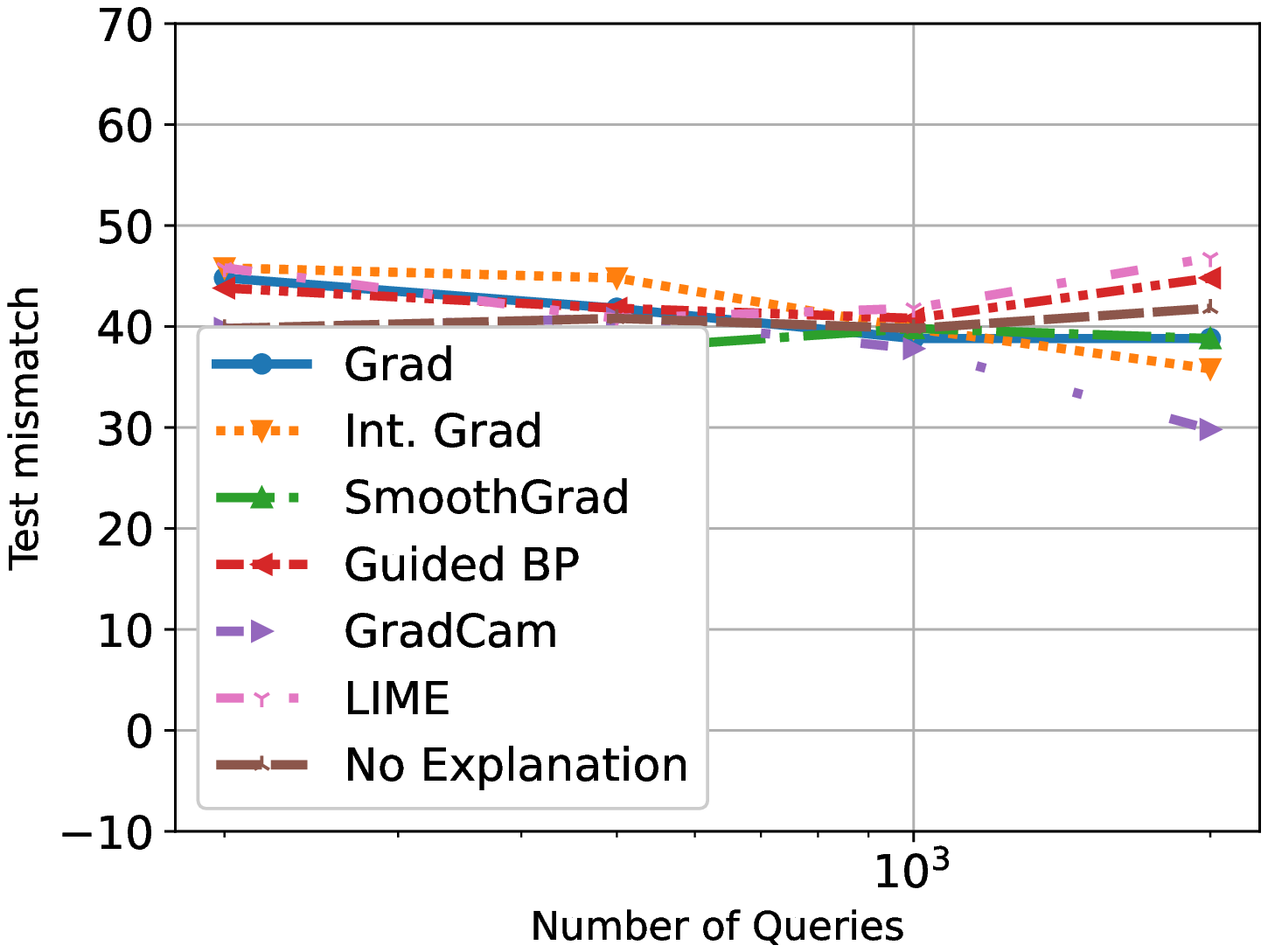} (c) Covid \centering  \par
    \endminipage\hfill
    \minipage{0.24\textwidth}
        \includegraphics[width=\linewidth]{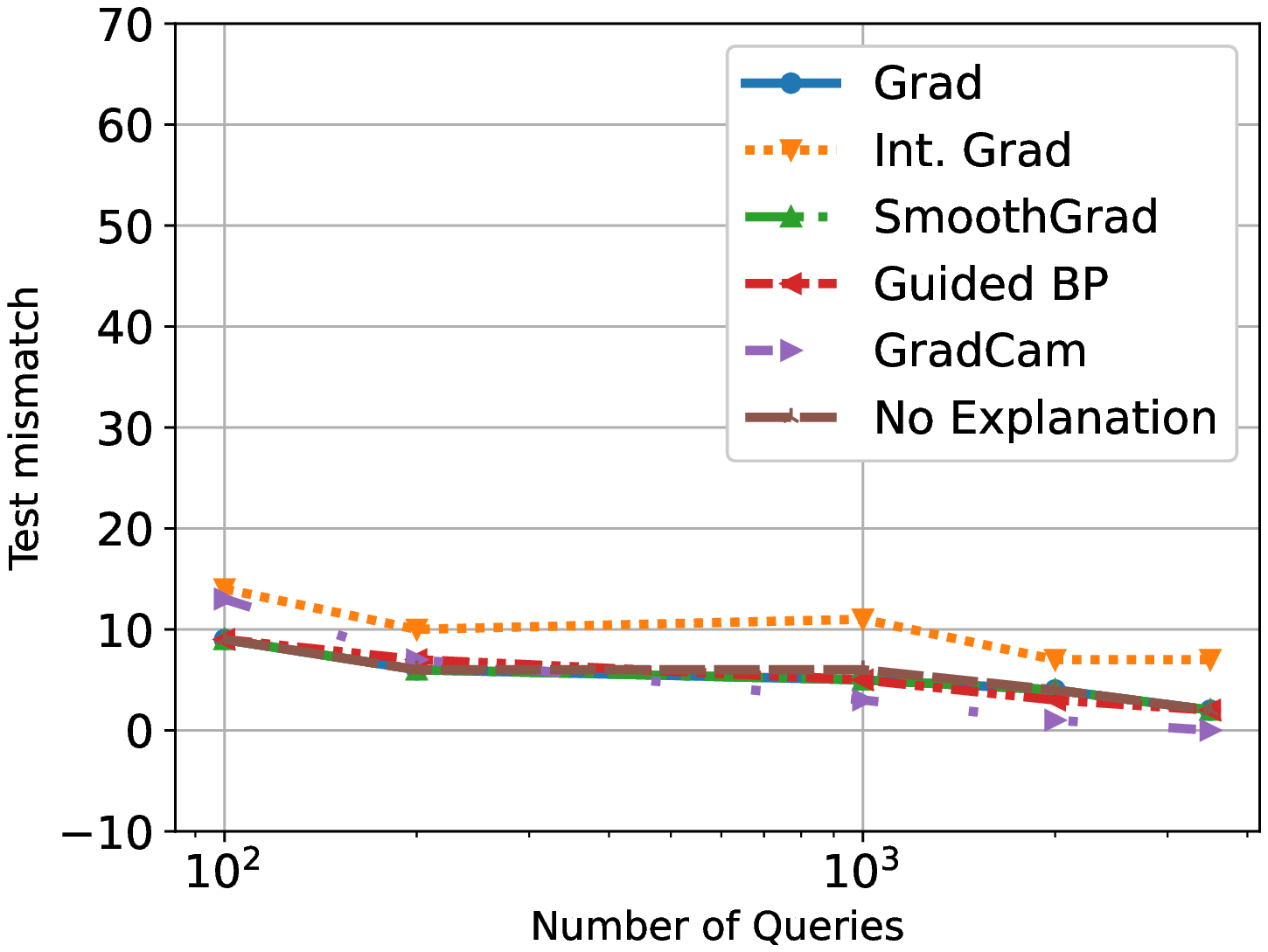} (d) HAR \centering \par
    \endminipage\hfill

		\caption{$R_{\text{test}}$ on various dataset. We show the performance of 6 various explanations and the baseline (without explanations).}
	\label{fig:MEA-cifar10}
\end{figure}

\subsubsection{MEA results}

Fig. \ref{fig:MEA-cifar10} (a) shows $R_{\text{test}}$ with respect to various of query budgets and explanation mechanisms on CIFAR-10 dataset. First, we can observe that with the help of the explanation matching mechanism, the adversary can have the advantage of producing an extracted model with higher accuracy. For instance, to produce an extracted model with 25 test mismatches, the without-explanation baseline consumes $10^4$ queries, while the Gradient and Guided Backprop gradient matching mechanism can have an 84\% query reduction. In this case, to perform the MEA attack on Machine learning as a service (MLaaS) using query API, the monetary costs are at $\approx$\$0.25 per query. \footnote{\url{https://www.clarifai.com/pricing}} Then, using the explanation-matching mechanism can reduce the cost by \$2100.


For the MNIST dataset, we can also have a similar observation where explanations of various forms can help reduce the test mismatch $R_{\text{test}}$ in Fig. \ref{fig:MEA-cifar10} (b). Among them, LIME, Guided Backprop, and Smooth Grad have the highest improvement using various query budgets. For instance, LIME can produce a task-accuracy equivalent model using $10^3$ queries, which has about 90\% costs reduction compared to the without-explanation baseline.


For the Covid dataset, the improvement is highly modest, as shown in Fig. \ref{fig:MEA-cifar10} (c). The baseline approach has about $R_{\text{test}}=40$. And most of the explanation-matching schemes fluctuate around the baseline with various query budgets. We also note that Gradient, LIME, and Guided Backprop have a clear increase in test mismatch, i.e., using the explanation-matching mechanism decreases the MEA performance. However, one clear exception is GradCam. Using 2000 queries, GradCam reduces $R_{\text{test}}$ from 42 to 30, where it shows around 25\% improvement.


Fig. \ref{fig:MEA-cifar10} (d) shows the MEA results on the HAR dataset. Similarly, we can observe that matching explanations do not improve the MEA performance using various queries in most cases. The only exception is GradCam, where the performance improves marginally. 

\subsubsection{Mean-Square-Value (MSV) of explanations as a strong indication}
Here we further investigate the reason why explanation matching has a clear improvement on the CIFAR-10 and MNIST dataset, while the improvement is relatively modest on the other two. Here we plot the Mean-Square-Value (MSV) with each explanation:

\begin{equation}
    MSV = \frac{1}{n} \sum_{i=1}^{n}E_i^2
\end{equation}

where $E \in \mathbb{R}^n$ is the explanation generated by the model. The MSV calculated the $L_2$-norm of $E$ normalized by its dimension. Table. \ref{table:MSV} shows the averaged MSV of different explanations of 1000 samples from each dataset. We can observe from Table. \ref{table:MSV} that the MSV has a strong correlation with the improvement made by each explanation method: a higher MSV can produce a larger advantage of performing MEA. We omit the value of LIME here since we does not rely on explanation matching but prediction matching to perform MEA.

First, the values of MSV of HAR and Covid are orders of magnitude smaller than those of CIFAR-10 and MNIST, which means that the explanations of HAR and Covid have limited information to be extracted. And matching the explanations does not produce a clear improvement in the performance of MEA. The relatively small gradient is caused by the overfitting of the model in the two datasets because instances from the training distribution will cause a small loss for the model and render a smaller gradient. On the other hand, the MSVs of explanations in the CIFAR-10 and MNIST datasets are consistently large, and we observe a more significant improvement of MEA. Besides, we also notice that the MSV of GradCam is consistently larger among all the datasets. And we notice a clear improvement in the attack performance produced by explanation matching using GradCam, which is in line with a larger MSV of GradCam.

\begin{table}[h]
	\centering
	\begin{adjustbox}{max width=0.43\textwidth}
		\begin{tabular}{c|c|c|c|c|c}
			\toprule
            Dataset & Grad & Int. Grad & Smooth Grad & Grad-Cam & Guided Backprop\\
            \midrule
            CIFAR-10 & 7.4e-1 & 7.8e-2 & 5.1e-1 & 5.2e-1 & 7.1e-1 \\
            \midrule
            MNIST & 7.5e-2 & 3.4e-2 & 7.3e-2 & 7.2e-2 & 7.4e-2 \\
            \midrule
            HAR & 7.0e-10 & 2.1e-4 & 7.0e-10 & 2.1e-1 & 7.8e-10 \\
            \midrule
            Covid & 2.2e-3 & 2.0e-3 & 9.0e-4 & 4.7e-2 & 1.5e-5\\
			\midrule
			\bottomrule
		\end{tabular}
	\end{adjustbox}
	\caption{Mean-Square-Value (MSV) of each explanation. The value is averaged over 1000 samples.}
	\label{table:MSV}
	\vspace{-3em}
\end{table}

\section{Discussion}
Our empirical analysis highlight that gradient-based explanations such as Gradient, SmoothGrad, and IntegratedGrad, allow an adversary to better estimate gradients in black-box settings. Therefore, attacks can be executed with lower number of queries (around $10\times$ for hard and soft label EA) making them efficient and increasing their feasibility in practice. Even when explanations reveal a function of the gradients (e.g., absolute value of gradients, LIME), our proposed attacks show improvement in attack efficiency ($2-5\times$ for EA, and increased leakage of membership information). However, for MEA the gains observed are modest ($5-10\times$ for CIFAR-10 and MNIST, and lower for Covid and HAR). We categorize the risk of explanation release into three classes based on the increase they cause in attack efficiency: high ($5-10\times$), medium ($2-5\times$) and low ($\leq 2\times$). We hope this will help practitioners in understanding the risks associated with explanation release and allow them to make informed decisions about the explanation technique to use.

There are other attacks that can be included as part of the risk profile. While we focused on attacks at model deployment time, there could be additional risks associated with training time attacks~\cite{shap_poisoning} as well. We emphasize that the goal of our study was not to be exhaustive but to initiate a systematic exploration of the interaction between explanations and privacy and security attacks and understand how different types of explanations effect risk. While our results indicate that explanations do represent an information-rich side-channel that can be exploited for improving attack efficiency, there is still work that needs to be done to make these attacks practical, especially in the access-constrained black-box settings.
\section{Conclusion}
In this paper, we proposed several attacks that leveraged post-hoc explanations to significantly improve attack efficiency of both evasion and membership inference attacks. The improvement in model extraction attack was modest, and we empirically analyzed the  performance to be strongly connected with the mean square value of the explanation. Finally, we categorized the risk level associated with different explanations and attacks. We hope this can serve as a guide for practitioners to identify the right type of explanations depending on the sensitivity of their tasks.



\section*{Acknowledgments}
This should be a simple paragraph before the References to thank those individuals and institutions who have supported your work on this article.

{\appendix
\section*{EGTA: A variance reduced approach}\label{sec:variance}
In this section, we theoretically quantify the advantage of EGTA. On one hand, EGTA is a biased estimator for the true gradient. On the other hand, as it relies on the transfer prior explanation $\boldsymbol e$, it can effectively reduce the variance of the estimation. 

Let $f(\boldsymbol x)$ be the objective function, i.e., the predicted probability of the target class given by the black-box machine learning model. Let $\boldsymbol e$ be the transfer prior explanation and we also let its norm $||\boldsymbol e||=1$. Let $\mathbf{u}_b$ be the search direction and $\mathbf{u}_b = \sqrt{\lambda} \mathbf{u} + \sqrt{1-\lambda} \mathbf{e} = \hat{\mathbf{u}} + \hat{\mathbf{e}}$ and $\mathbf{u} \sim \mathcal N(\mathbf{0}, \epsilon\mathbf{I})$. Then the estimation gradient using natural evolution strategy becomes:

\begin{equation}\label{nes_suppl}
    \widetilde{\nabla f(\mathbf{x})} = \frac{1}{2\delta B} \sum_{b=1}^B (f(\mathbf{x}+\delta\mathbf{u}_b)-f(\mathbf{x}-\delta\mathbf{u}_b))\mathbf{u}_b 
\end{equation}

W.l.o.g, we assume $\delta$ and $\epsilon$ equal to 1. For the purposes of analysis, suppose $\nabla f$ exists. We can approximate the function in the local neighborhood of $\boldsymbol x$ using a second order Taylor approximation: $f(\boldsymbol x + \mathbf{u}_b) \approx f(\boldsymbol x) + \mathbf{u}_b^T\nabla f(\boldsymbol x) + \frac{1}{2}\mathbf{u}_b^T\nabla^2 f(\boldsymbol x)\mathbf{u}_b$. Substituting this expression into \eqref{nes_suppl}, we can see our estimate $\widetilde{\nabla f(\mathbf{x})}$ equal to:

\begin{equation}
    \widetilde{\nabla f(\mathbf{x})} = \frac{1}{B}\sum_{b=1}^B (\mathbf{u}_b\mathbf{u}_b^T)\nabla f(\mathbf{x})
\end{equation}

Then the expectation of $\mathbb{E}[\widetilde{\nabla f(\mathbf{x})}]$ equals:

\begin{equation}\label{eqn:mean}
    \mathbb{E}[\widetilde{\nabla f(\mathbf{x})}] = (\hat \Sigma+\hat G)\nabla f(\mathbf{x})
\end{equation}

Hence bias equals:

\begin{align}\label{eqn:bias}
		\text{bias} &= ||\mathbb{E}[\widetilde{\nabla f(\mathbf{x})}] - \nabla f(\mathbf{x})||_2^2\\
		  & = (1-\lambda)^2 \nabla f(\mathbf{x})^T(I-\boldsymbol{e} \boldsymbol{e}^T)^2\nabla f(\mathbf{x}) \\
		  & = (1-\lambda)^2(|| \nabla f(\mathbf{x})||_2^2 - ( \nabla f(\mathbf{x})^T\boldsymbol{e})^2) \\
		  & = (1-\lambda)^2 || \nabla f(\mathbf{x})||_2^2(1-\text{cos}(\theta)^2)
\end{align}

where $\text{cos}(\theta)$ is the cosine similarity between $\boldsymbol{e}$ and $\nabla f(\mathbf{x})$. When cosine similarity is high, the bias of the estimation is low.

Next we use the total variance, i.e., $\text{tr}(\text{Var}(\widetilde{\nabla f(\mathbf{x})})$, to quantify the variance of our estimator. Since $\boldsymbol{u}_i$ and $\boldsymbol{u}_j$ are independent random variables and have the same variance:.

\begin{equation}  \label{eqn:tv}
    \text{Var}(\widetilde{\nabla f(\mathbf{x})}) = \frac{1}{B} \text{Var}((\mathbf{u}_b\mathbf{u}_b^T)\nabla f(\mathbf{x}))
\end{equation}

\begin{align*} 
		\text{total variance} &= \text{tr}(\text{Var}(\widetilde{\nabla f(\mathbf{x})}) \\
		& = \text{tr}(\mathbb{E}[\widetilde{\nabla f(\mathbf{x})} \widetilde{\nabla f(\mathbf{x})} ^T]) - \text{tr}(\mathbb{E}[\widetilde{\nabla f(\mathbf{x})}]\mathbb{E}[\widetilde{\nabla f(\mathbf{x})}]^T)\\
		& = \frac{1}{B}(\nabla f(\mathbf{x})^T \mathbb{E}[\boldsymbol{u}_b \boldsymbol{u}_b^T \boldsymbol{u}_b \boldsymbol{u}_b^T] \nabla f(\mathbf{x}) - \\ 
		& \mathbb{E}[\widetilde{\nabla f(\mathbf{x})}]^T\mathbb{E}[\widetilde{\nabla f(\mathbf{x})}])
\end{align*}

Next, we will quantify the term $\mathbb{E}[\boldsymbol{u}_b \boldsymbol{u}_b^T \boldsymbol{u}_b \boldsymbol{u}_b^T]$. By Isserlis’ theorem, the odd order moment  $\mathbb{E}[\hat{u}_i \hat{u}_j \hat{u}_k]$ and $\mathbb{E}[\hat{u}_i]$ equal to 0:

\begin{align*}\label{eqnn:uuuu}
    \mathbb{E}[\boldsymbol{u}_b \boldsymbol{u}_b^T \boldsymbol{u}_b \boldsymbol{u}_b^T] &= \mathbb{E}[(\hat{\mathbf{u}} + \hat{\mathbf{e}})(\hat{\mathbf{u}} + \hat{\mathbf{e}})^T(\hat{\mathbf{u}} + \hat{\mathbf{e}})(\hat{\mathbf{u}} + \hat{\mathbf{e}})^T] \\
    &= \mathbb{E}[\hat{\mathbf{u}}\hat{\mathbf{u}}^T\hat{\mathbf{u}}\hat{\mathbf{u}}^T] + 2 \mathbb{E}[\hat{\mathbf{e}}\hat{\mathbf{e}}^T\hat{\mathbf{u}}\hat{\mathbf{u}}^T] +  \\
    & 4 \mathbb{E}[\hat{\mathbf{e}} \hat{\mathbf{u}}^T \hat{\mathbf{e}} \hat{\mathbf{u}}^T] + \mathbb{E}[\hat{\mathbf{e}}\hat{\mathbf{e}}^T\hat{\mathbf{e}}\hat{\mathbf{e}}^T]
\end{align*}

We can observe that $\mathbb{E}[\hat{\mathbf{e}} \hat{\mathbf{u}}^T \hat{\mathbf{e}} \hat{\mathbf{u}}^T] = \hat{G}\hat{\Sigma}$, which is by noticing that $\mathbb{E}[\hat{\mathbf{e}} \hat{\mathbf{u}}^T \hat{\mathbf{e}} \hat{\mathbf{u}}^T]_{i,j} = \hat{e}_i\hat{e}_j \mathbb{E}[\hat{u}_j^2]$. Similarly, we can also observe that $\mathbb{E}[\hat{\mathbf{e}}\hat{\mathbf{e}}^T\hat{\mathbf{u}}\hat{\mathbf{u}}^T]= \hat{G}\hat{\Sigma}$.

By applying the Isserlis’ theorem again, we can simplify $\mathbb{E}[\hat{\mathbf{u}}\hat{\mathbf{u}}^T\hat{\mathbf{u}}\hat{\mathbf{u}}^T] = \text{tr}(\hat{\Sigma})\hat{\Sigma}+2\hat{\Sigma}^2$. Therefore, \eqref{eqnn:uuuu} equals:

\begin{align}
    \mathbb{E}[\boldsymbol{u}_b \boldsymbol{u}_b^T \boldsymbol{u}_b \boldsymbol{u}_b^T] &= \text{tr}(\hat{\Sigma})\hat{\Sigma}+2\hat{\Sigma}^2 + 6\hat{G}\hat{\Sigma} + \hat{G}^2
\end{align}

Then \eqref{eqn:tv} equals (by \eqref{eqnn:uuuu} and \eqref{eqn:mean}):

\begin{align*}
		\text{total variance} &= \frac{1}{B} (\nabla f(\mathbf{x})^T \mathbb{E}[\boldsymbol{u}_b \boldsymbol{u}_b^T \boldsymbol{u}_b \boldsymbol{u}_b^T] \nabla f(\mathbf{x}) - \\
		& \mathbb{E}[\widetilde{\nabla f(\mathbf{x})}]^T\mathbb{E}[\widetilde{\nabla f(\mathbf{x})}]) \\
		&= \frac{1}{B}(\nabla f(\mathbf{x})^T \mathbb{E}[\boldsymbol{u}_b \boldsymbol{u}_b^T \boldsymbol{u}_b \boldsymbol{u}_b^T] \nabla f(\mathbf{x}) - \\ 
		& \nabla f(\mathbf{x})^T(\hat \Sigma+\hat G)(\hat \Sigma+\hat G)\nabla f(\mathbf{x})) \\ 
		& = \frac{1}{B}(\nabla f(\mathbf{x})^T (\text{tr}(\hat{\Sigma})\hat{\Sigma}+\hat{\Sigma}^2 + 4\hat{G}\hat{\Sigma}) \nabla f(\mathbf{x}))
\end{align*}

Since $\hat \Sigma=\lambda I$ and $\hat G = (1-\lambda)\boldsymbol{e} \boldsymbol{e}^T$, then the above equals:

\begin{align*}
    \text{total variance} &= \frac{1}{B}(\nabla f(\mathbf{x})^T (\lambda^2(n+1)I+ \\
    & 4(1-\lambda)\lambda \boldsymbol{e} \boldsymbol{e}^T ) \nabla f(\mathbf{x})) \\
    & = \frac{1}{B}(\lambda^2(n+1)||\nabla f(\mathbf{x})||_2^2 + \\
    & 4(1-\lambda)\lambda (\nabla f(\mathbf{x})^T \boldsymbol{e})^2)
\end{align*}

Especially,

\begin{equation*}
    \text{total variance} = \left\{\begin{array}{ll}
        \frac{n+1}{B}||\nabla f(\mathbf{x})||_2^2, & \text{when } \lambda=1\\
        0, & \lambda=0
        \end{array}
        \right.
\end{equation*}

$(n+1)||\nabla f(\mathbf{x})||_2^2 \gg 4 (\nabla f(\mathbf{x})^T \boldsymbol{e})^2 $ since $n$ is typically larger than 1 for high dimensional data such as images and $||\nabla f(\mathbf{x})||_2^2 \geq (\nabla f(\mathbf{x})^T \boldsymbol{e})^2$. Therefore, when $\lambda$ is close to 0, i.e., the search direction is more relying on the prior explanation, the variance will be much lower than the case where no search direction is provided.

\section*{EGSA: the optimal approximation}\label{sec:opt_approx}
Similarly, instead of changing the mean of the search distribution using the transfer prior, \cite{cheng2019improving} shape the covariance of the distrbution by $\mathbf{u}_b = \sqrt{\lambda} \mathbf{u} + \sqrt{1-\lambda}u' \mathbf{e}$ where $u'\sim\mathcal{N}(0,1)$ to trades off variance between the full parameter space and the prior space. In this way, changing $\mathbf{u}_b$ in the above manner is theoretically equivalent to adapting the covariance matrix of the sampling distribution \cite{cheng2019improving}:

\begin{equation}\label{eqn: sigma}
    \mathbf{\Sigma} = \lambda \epsilon \mathbf{I} + (1-\lambda) \mathbf{e}\mathbf{e}^T
\end{equation}

In this attack, we proposed EGSA by element-wise scaling:

\begin{equation}\label{eq:transfer_sample}
    \mathbf{u}_b = \sqrt{\lambda} \mathbf{u} + \sqrt{1-\lambda} \mathbf{|e|} \odot \mathbf{u}
\end{equation} 

Therefore, the covariance matrix of the search distribution as follows:
\begin{equation}\label{eq:transfer}
    \mathbf{\Sigma^*} = \lambda \epsilon \mathbf{I} + (1-\lambda) \text{diag}(\mathbf{|e|}\odot\mathbf{|e|})
\end{equation}

The benefit of shaping the covariance matrix as $\mathbf{\Sigma^*}$ is that it provides us a substitution for $\mathbf{\Sigma}$ when the off-diagonal terms $\sigma_{i,j}$ are not available. In the followings, we show that $\mathbf{\Sigma^*}$ is the minimizer of the \textit{KL divergence} between the distribution $\mathcal N(\mathbf{0}, \mathbf{\Sigma})$ and the distribution $\mathcal N(\mathbf{0}, \mathbf{\Sigma'})$ for any $\mathbf{\Sigma'} \in \mathbb{D}=  \{\text{diag}(\sigma_{11},...\sigma_{NN}) | \sigma_{ii} \geq 0 \text{ for }\forall i \in [1,...,N]\}$. 

\begin{Proposition}
Denote $\mathcal{D}_1 = \mathcal N(\mathbf{0}, \mathbf{\Sigma})$ where $\mathbf{\Sigma}$ is shown in Eqn. \eqref{eqn: sigma}. For any Gaussian distribution $\mathcal{D}_2 \in \{\mathcal N(\mathbf{0}, \mathbf{\Sigma}')|\mathbf{\Sigma}'\in \mathbb{D}\}$, $\mathbf{\Sigma^*}$ minimizes the \textit{KL divergence} between $\mathcal{D}_1$ and $\mathcal{D}_2$.
\end{Proposition}

\begin{proof}

The \textit{KL divergence} is $D_{KL}(\mathcal{D}_1||\mathcal{D}_2)=\frac{1}{2}[\text{log}\frac{|\mathbf{\Sigma}_2|}{|\mathbf{\Sigma}_1|} - n +\text{tr}(\mathbf{\Sigma}_2^{-1}\mathbf{\Sigma}_1)+(\boldsymbol{\mu}_1-\boldsymbol{\mu}_2)^T\mathbf{\Sigma}_2^{-1}(\boldsymbol{\mu}_1-\boldsymbol{\mu}_2)]$. From assumption, we know that both $\mathcal{D}_1$ and $\mathcal{D}_2$ have zero mean. Then $D_{KL}(\mathcal{D}_1||\mathcal{D}_2)=\frac{1}{2}[\text{log}\frac{|\mathbf{\Sigma}_2|}{|\mathbf{\Sigma}_1|} - n +\text{tr}(\mathbf{\Sigma}_2^{-1}\mathbf{\Sigma}_1)]$.


We can formulate the optimization problem as follows:

	\begin{subequations}\label{L2_loss}
		\begin{alignat}{2}
		&\underset{\mathbf\Sigma'}{\text{minimize}}& \qquad & D_{KL}(\mathcal{D}_1||\mathcal{D}_2)\\
		&\text{s.t.} & & \mathbf \Sigma' \in \mathbb{D} \\
		&& & \mathbf{\Sigma} = \lambda \epsilon \mathbf{I} + (1-\lambda) \mathbf{e}\mathbf{e}^T
		\end{alignat}
	\end{subequations}


Further simplification leads to, $D_{KL}(\mathcal{D}_1||\mathcal{D}_2)=\frac{1}{2}[\sum_{i=1}^n\text{log}\sigma'_{ii}-\sum_{i=1}^n\text{log}\sigma_{ii}-n+\sum_{i=1}^n{\sigma'_{ii}}^{-1}\sigma_{ii}]$. Then the partial derivative with respect to each $\sigma'_{ii}$ is $\frac{\partial}{\partial \sigma'_{ii} }D_{KL}(\mathcal{D}_1||\mathcal{D}_2) ={\sigma'_{ii}}^{-1}-{\sigma'_{ii}}^{-2}\sigma_{ii}$. By setting the derivative to $0$, we can find the minimizer of each $\sigma'_{ii}=\sigma_{ii}$. This effectively tells us that the $\mathcal{D^*}=\mathcal{N}(\mathbf{0}, \mathbf{\Sigma^*})$ is a close approximation of the distribution $\mathcal{D}_1=\mathcal{N}(\mathbf{0}, \mathbf{\Sigma})$.
\end{proof}

The above discussion indicates that given the explanation $\mathbf E$, the attacker can incorporate it as the transfer prior through Eqn. \eqref{eq:transfer} since this will render a closer \textit{KL divergence} than any other distribution with a diagonal matrix. 

\section*{Dataset information}\label{sec:dataset}
\noindent \textbf{CIFAR-10. \cite{krizhevsky2009learning}} This is a benchmark dataset used to evaluate image recognition algorithms. The image is composed of 32×32 color images in 10 classes, with 6,000 images per class. 

\noindent \textbf{Purchases. \cite{shokri2017membership}} This is based on Kaggle’s “acquire valued shoppers” challenge dataset that contains shopping histories for several thousand consumers. Each individual record is a binary vector of 600 dimensions and there are 100 classes in total.

\noindent \textbf{Texas. \cite{shokri2017membership}} This dataset is based on the Hospital Discharge Data public use files with information about inpatients stays in several health facilities. Each individual record is a binary vector of 6170 dimensions and there are 100 classes in total.

\noindent \textbf{Human Activity Recognition (HAR). \cite{li2021units, roggen2010collecting}} This dataset in used for human locomotion recognition uses wearable motion sensors with inertial measurement units (IMU). Following \cite{li2021units}, each data example is a $45 \times 256$ matrix and there are 4 classes in total.

\noindent \textbf{Covid. \cite{rahman_2021}} This is a database of chest X-ray images to aid Covid-19 pneumonia diagnose, including 4 classes of COVID-19, normal, Viral Pneumonia, and Lung Opacity.

\section*{Acknowledgement}
The research presented in this paper is supported in part by the CONIX Research Center, one of
six centers in JUMP, a Semiconductor Research Corporation (SRC) program sponsored by DARPA,
the U.S. Army Research Laboratory and the U.K. Ministry of Defence under Agreement Number
W911NF-16-3-0001 and the National Science Foundation (NSF) under award \# CNS-1822935, and
by the National Institutes of Health (NIH) award \# P41EB028242 for the mDOT Center. The views
and conclusions contained in this document are those of the authors and should not be interpreted as
representing the official policies, either expressed or implied, of the NSF, the NIH, the U.S. Army
Research Laboratory, the U.S. Government, the U.K. Ministry of Defence or the U.K. Government.
The U.S. and U.K. Governments are authorized to reproduce and distribute reprints for Government
purposes notwithstanding any copyright notation hereon. This work has been submitted to the IEEE for possible publication. Copyright may be transferred without notice, after which this version may no longer be accessible.

}


\bibliographystyle{IEEEtran}
\bibliography{references}


 





\end{document}